\documentclass[10pt,twocolumn,letterpaper]{article}

\usepackage[final]{cvpr}      %

\usepackage[dvipsnames]{xcolor}

\usepackage{graphicx}
\usepackage{booktabs}
\usepackage{xcolor,colortbl}
\usepackage{caption}
\usepackage{subcaption}
\usepackage{amsmath}
\usepackage[bb=dsserif]{mathalpha}
\usepackage{bm}
\usepackage{floatrow}
\usepackage{float}
\definecolor{cvprblue}{rgb}{0.21,0.49,0.74}
\usepackage[pagebackref,breaklinks,colorlinks,citecolor=cvprblue]{hyperref}

\def\paperID{00007} %
\def\confName{3DV\xspace}
\def\confYear{2025\xspace}

\title{RadSplat: Radiance Field-Informed Gaussian Splatting for Robust Real-Time Rendering with 900+ FPS}

\author{Michael Niemeyer \hspace{.5cm} Fabian Manhardt \hspace{.5cm} Marie-Julie Rakotosaona\\Michael Oechsle \hspace{.5cm}  Daniel Duckworth \hspace{.5cm} Rama Gosula\\Keisuke Tateno \hspace{.5cm} John Bates \hspace{.5cm} Dominik Kaeser \hspace{.5cm} Federico Tombari\\
Google\\
{\tt \url{m-niemeyer.github.io/radsplat}}
}

\newcommand{\bc}{\mathbf{c}}
\newcommand{\bd}{\mathbf{d}}

\newcommand{\bk}{\mathbf{k}}

\newcommand{\bp}{\mathbf{p}}
\newcommand{\bq}{\mathbf{q}}
\newcommand{\br}{\mathbf{r}}
\newcommand{\bs}{\mathbf{s}}

\newcommand{\bx}{\mathbf{x}}

\newcommand{\nR}{\mathbb{R}}

\newcommand{\cI}{\mathcal{I}}

\newcommand{\cK}{\mathcal{K}}
\newcommand{\cL}{\mathcal{L}}

\newcommand{\cP}{\mathcal{P}}

\newcommand{\cR}{\mathcal{R}}

\newcommand{\cU}{\mathcal{U}}

\makeatletter
\DeclareRobustCommand\onedot{\futurelet\@let@token\@onedot}
\def\@onedot{\ifx\@let@token.\else.\null\fi\xspace}
\def\eg{e.g\onedot} 
\def\ie{i.e\onedot}

\makeatother

\newcommand{\boldparagraph}[1]{\vspace{0.2cm}\noindent{\bf #1.}}

\newcommand{\secref}[1]{{Sec.\ \ref{#1}}}
\newcommand{\figref}[1]{{Fig.\ \ref{#1}}}
\newcommand{\tabref}[1]{{Tab.\ \ref{#1}}}

\newcommand{\zipnerf}{Zip-NeRF\xspace}

\begin{document}
\maketitle
\begin{abstract}
  Recent advances in view synthesis and real-time rendering have achieved photorealistic quality at impressive rendering speeds. While radiance field-based methods achieve state-of-the-art quality in challenging scenarios such as in-the-wild captures and large-scale scenes, they often suffer from excessively high compute requirements linked to volumetric rendering. Gaussian Splatting-based methods, on the other hand, rely on rasterization and naturally achieve real-time rendering but suffer from brittle optimization heuristics that underperform on more challenging scenes. In this work, we present RadSplat, a lightweight method for robust real-time rendering of complex scenes. Our main contributions are threefold. First, we use radiance fields as a prior and supervision signal for optimizing point-based scene representations, leading to improved quality and more robust optimization. Next, we develop a novel pruning technique reducing the overall point count while maintaining high quality, leading to smaller and more compact scene representations with faster inference speeds. Finally, we propose a novel test-time filtering approach that further accelerates rendering and allows to scale to larger, house-sized scenes. We find that our method enables state-of-the-art synthesis of complex captures at 900+ FPS.
\end{abstract}
    
\section{Introduction}
\label{sec:intro}

Neural fields~\cite{mescheder2019occupancynet, park2019deepsdf, chen2018implicit_decoder,xie2022neural} have emerged as one of the most popular representations for 3D vision due to their simple design, stable optimization, and state-of-the-art performance.
After their introduction in the context of 3D reconstruction~\cite{mescheder2019occupancynet, park2019deepsdf, chen2018implicit_decoder,xie2022neural}, neural fields have been widely adopted and set new standards in tasks such as
view synthesis~\cite{mildenhall2020nerf,barron2022mip,Barron2023zipnerf},
3D and 4D reconstruction~\cite{yariv2023baked,li2023neuralangelo,park2021hypernerf,niemeyer2019occupancy,park2021nerfies,peng2020convolutional},
and generative modeling~\cite{poole2022dreamfusion,niemeyer2021giraffe,schwarz2020graf,Lin2023magic3d, tsalicoglou2023textmesh}.

\begin{figure}
    \centering
    \includegraphics[width=\linewidth
    ]{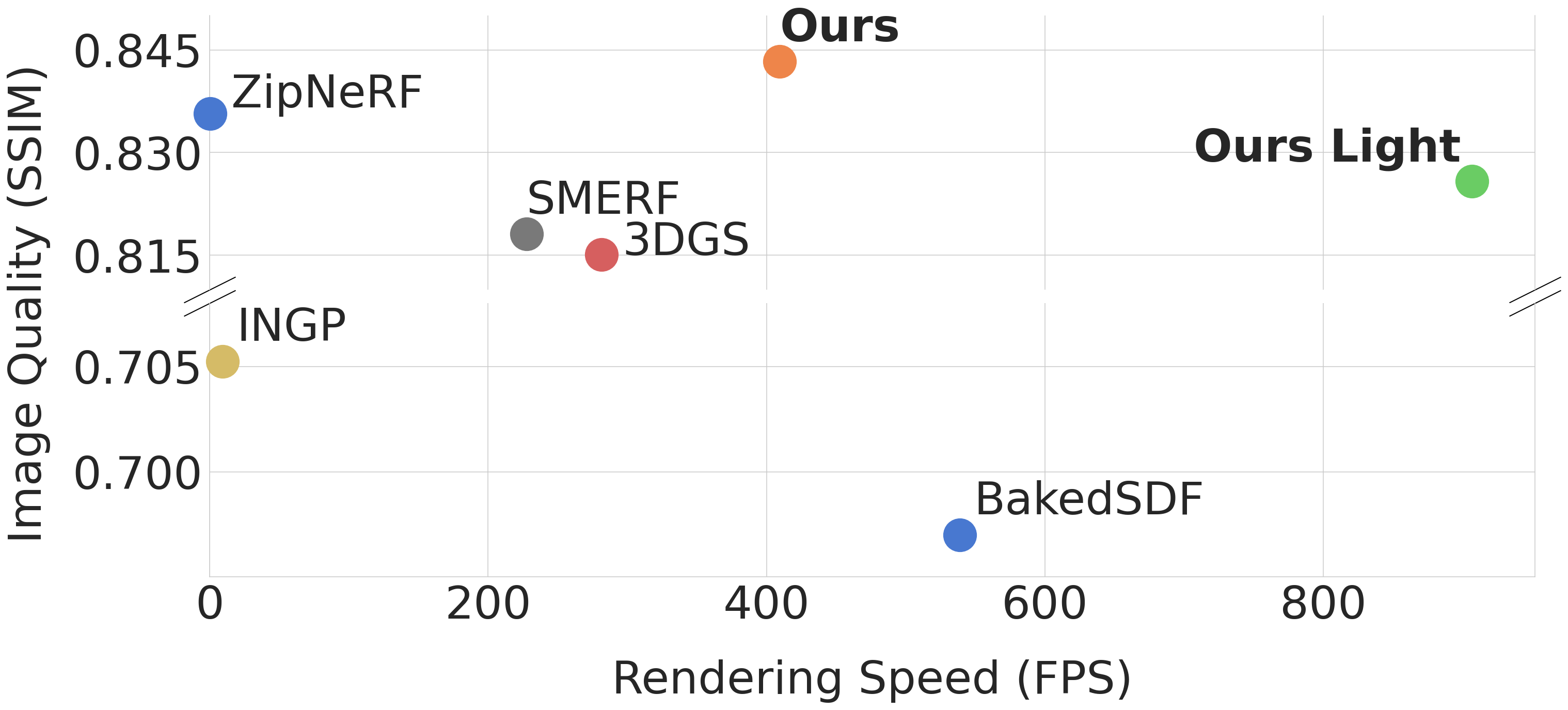}
    \vspace{-.6cm}
    \caption{
        \textbf{RadSplat.} By combining benefits of neural fields and point-based representations, we achieve state-of-the-art quality in view synthesis on mip-NeRF 360~\cite{barron2022mip} while rendering at 900+ frames per second (FPS), indicating a speed up of 3.6$\times$ over 3D Gaussian Splatting (3DGS)~\cite{kerbl20233d} and 3,000$\times$ over \zipnerf{}~\cite{Barron2023zipnerf}.
    }\label{fig:teaser}
\end{figure} 

While neural field methods have achieved unprecedented view synthesis quality even for challenging real-world captures~\cite{Barron2023zipnerf,mildenhall2020nerf,brualla2021wild}, most approaches are limited by the high compute costs of volumetric rendering.
In order to achieve real-time frame rates, recent works reduce network complexity~\cite{yu2021plenoxels,mueller2022instant},
cache intermediate outputs~\cite{reiser2023merf,duckworth2023smerf},
or extract 3D meshes~\cite{yariv2021volsdf,oechsle2021unisurf,yariv2023baked,rakotosaona2023nerfmeshing}.
Nevertheless, all methods trade reduced quality and increased storage costs for faster rendering, and are incapable of maintaining state-of-the-art quality in real-time -- the goal of this work (see~\figref{fig:teaser}).

\begin{figure}[ht]
\centering
\begin{subfigure}{.49\linewidth}
\includegraphics[width=\linewidth]{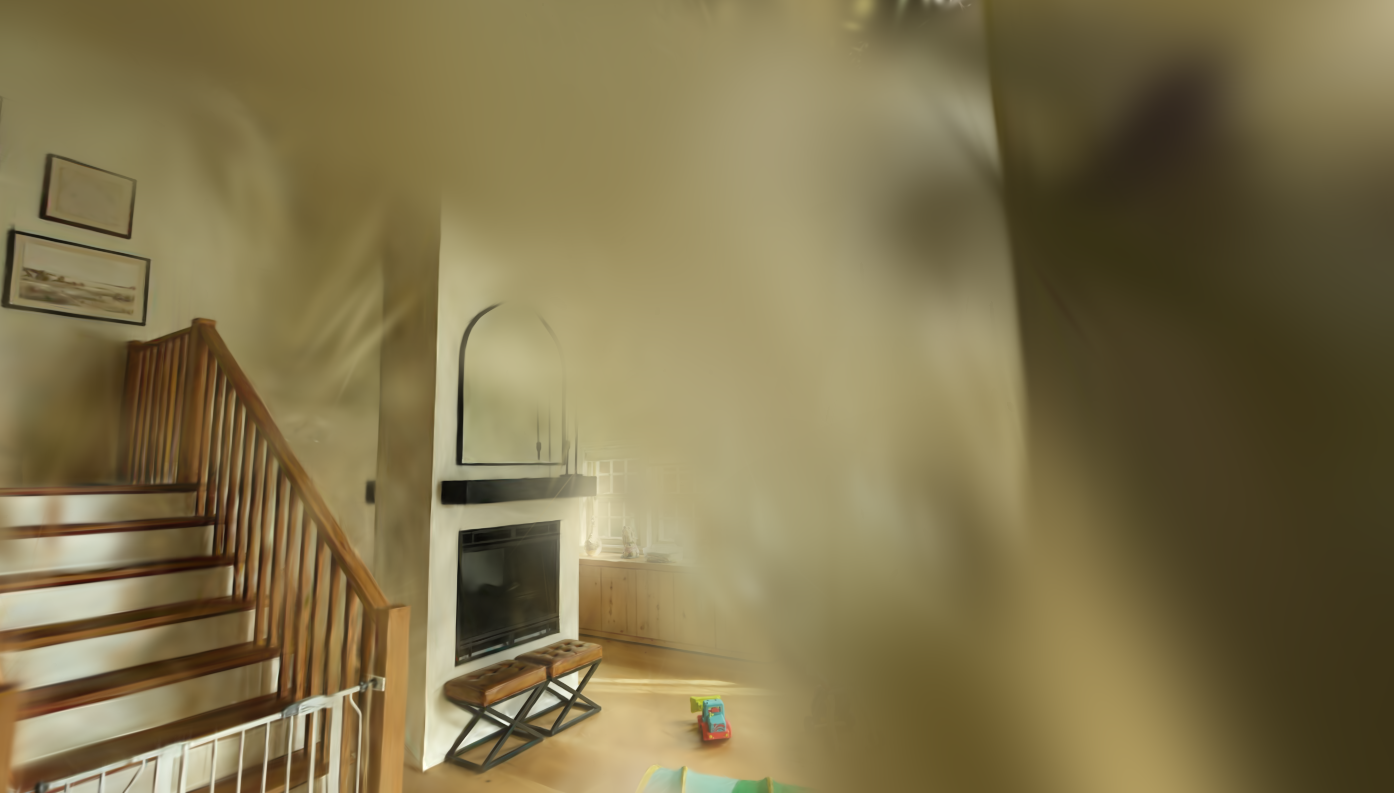}
\subcaption{3DGS.}\label{subfig:robust-a}
\end{subfigure}
\begin{subfigure}{.49\linewidth}
\includegraphics[width=\linewidth]{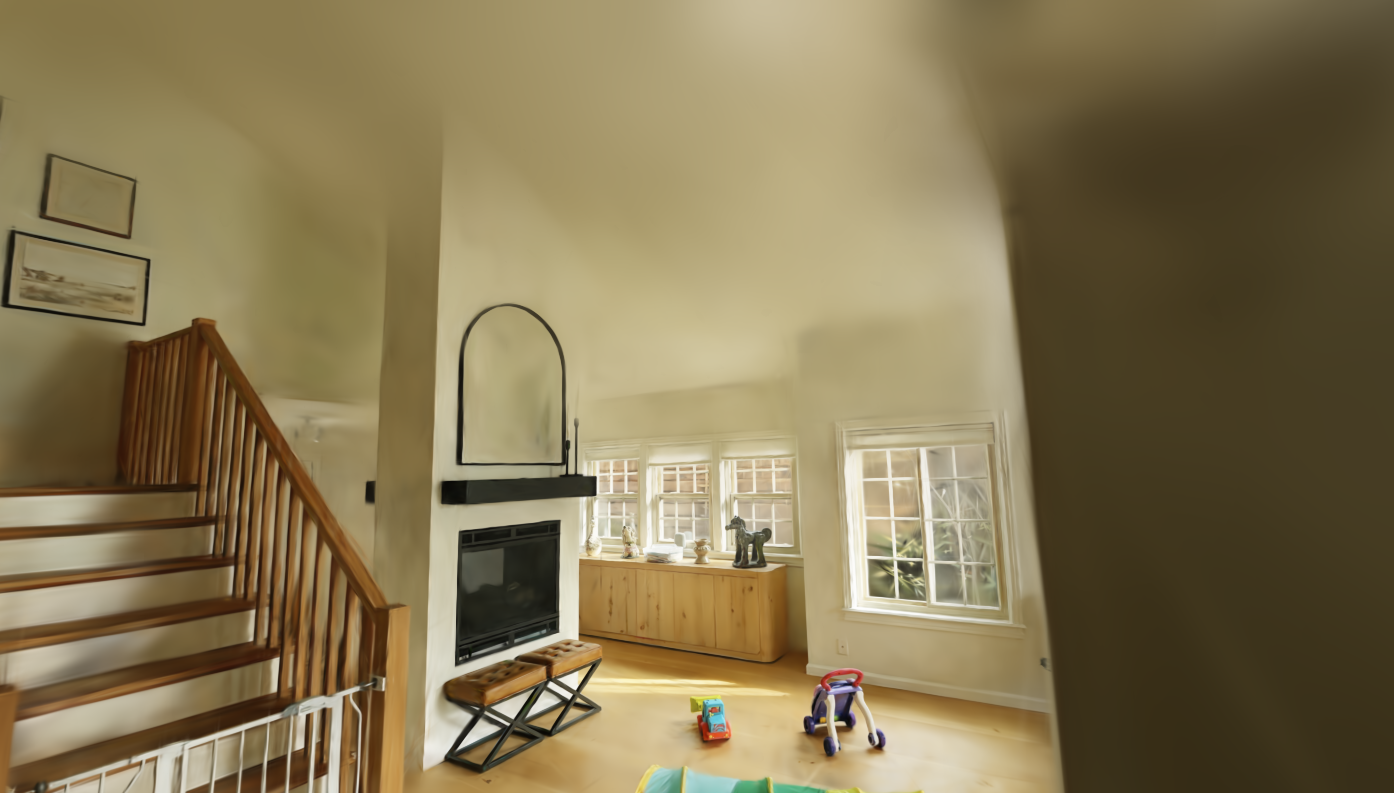}
\subcaption{3DGS with exposure module.}\label{subfig:robust-b}
\end{subfigure}
\begin{subfigure}{1.0\linewidth}
\includegraphics[width=.49\linewidth]{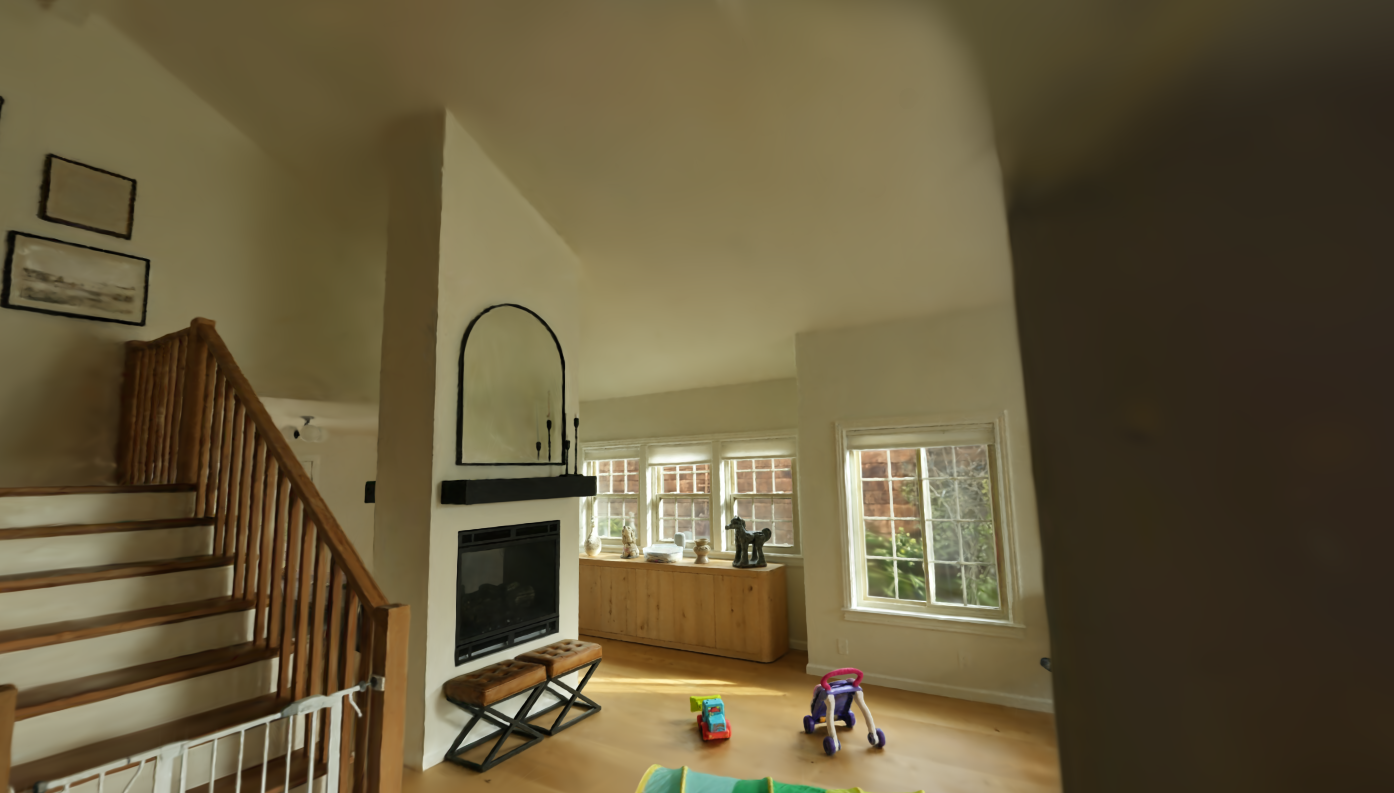}
\includegraphics[width=.49\linewidth]{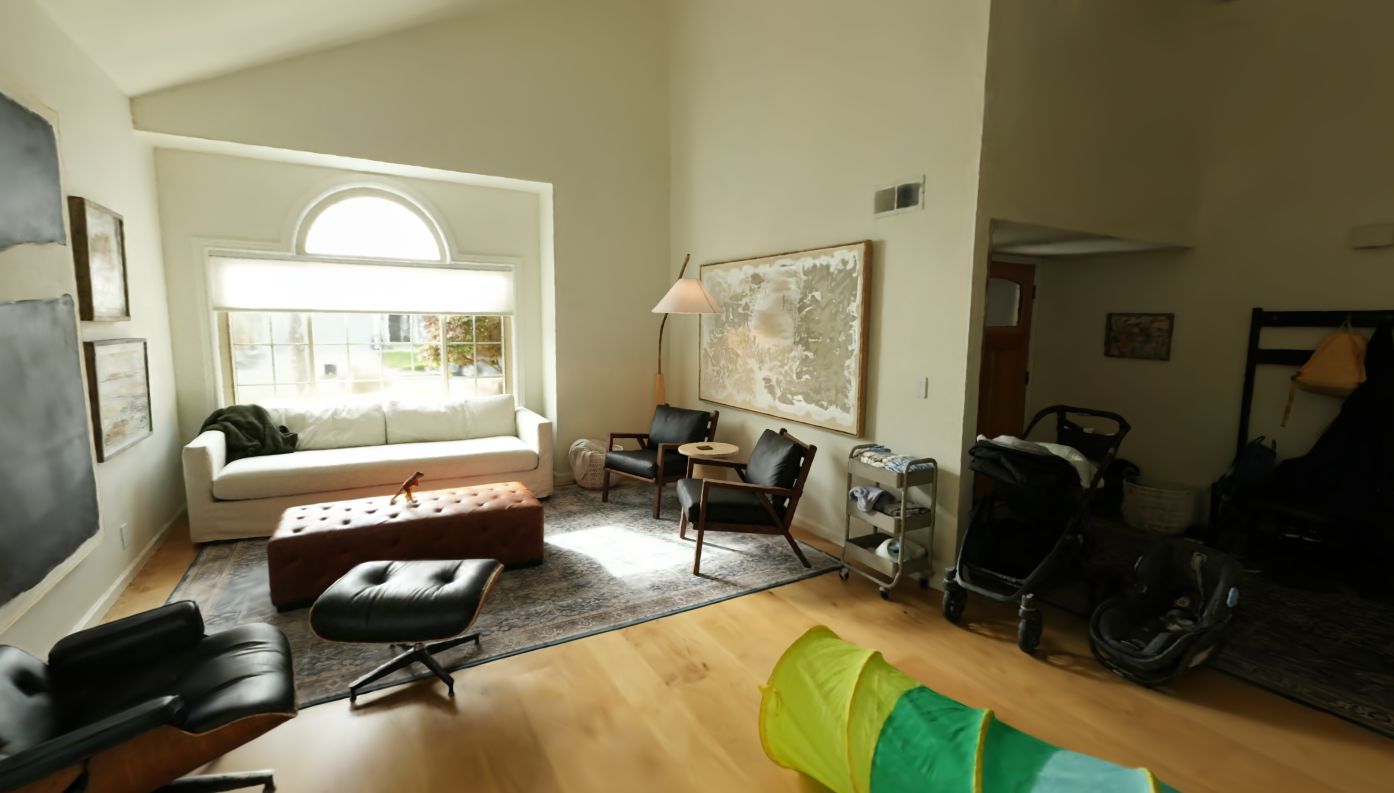}
\subcaption{RadSplat (Ours).}\label{subfig:robust-c}
\end{subfigure}    \vspace{-.6cm}

\caption{
    \textbf{Robust View Synthesis.} On complex scenes with lighting variations, 3D Gaussian Splatting (3DGS)~\cite{kerbl20233d} degrades (\ref{subfig:robust-a}). When equipped with exposure handling modules~\cite{kerbl20233d,duckworth2023smerf}, results improve but they still contain  artifacts and are overly smooth (\ref{subfig:robust-b}). In contrast, we
    achieve high quality even for challenging, large-scale captures (\ref{subfig:robust-c}) by integrating a robust radiance field as prior.
}
\label{fig:robustness}
\end{figure}

Recently, rasterization-based 3D Gaussian Splatting (3DGS)~\cite{kerbl20233d} has emerged as a natural alternative to neural fields.
The representation admits real-time frame rates with view synthesis quality rivaling the state-of-the-art in neural fields.
3DGS, however, suffers from a challenging optimization landscape and an unbounded model size.
The number of Gaussian primitives is not known a priori, and carefully-tuned merging, splitting, and pruning heuristics are required to achieve satisfactory results.
The brittleness of these heuristics become particularly evident in large scenes where phenomena such as exposure variation, motion blur, and moving objects are unavoidable (see~\figref{fig:robustness}).
An increasing number of primitives further leads to a potentially-unmanageable memory footprint and reduced rendering speed, strongly limiting model quality for larger scenes.

In this work, we present RadSplat, a lightweight method for robust real-time rendering of complex real-world scenes.
Our method achieves smaller model sizes and faster rendering than 3DGS while strongly exceeding reconstruction quality.
Our key idea is to combine the stable optimization and quality of neural fields to act as a prior and supervision signal for the optimization of point-based scene representations.
We further introduce a novel pruning procedure and test-time visibility rendering strategies to significantly reduce memory usage and increase rendering speed without a corresponding loss in quality.
In summary, our contributions are as follows:  
\vspace{.1cm}
\begin{enumerate}
    \item The use of radiance fields as a prior and to handle the complexity of real-world data when optimizing point-based 3DGS representations. 
    \item A novel pruning strategy that reduces the number of Gaussian primitives by up to 10x whilst improving quality and rendering speed.
    \item A novel post-processing step enabling viewpoint-based filtering, further accelerating rendering speed without any reduction in quality. 
\end{enumerate}
\vspace{.1cm}
Our method exhibits state-of-the-art reconstruction quality on both medium and large scenes, with PSNR up to 1.87 dB higher than 3DGS and SSIM exceeding \zipnerf{}, the current state-of-the-art in offline view synthesis (see~\figref{fig:teaser}).
At the same time, our method renders up to 907 frames per second, over 3.6$\times$ faster than 3DGS~\cite{kerbl20233d} and more than 3,000$\times$ faster than \zipnerf{}~\cite{Barron2023zipnerf}.

\section{Related Work}
\label{sec:relwork}

\boldparagraph{Neural Fields}
Since their introduction in the context of 3D reconstruction~\cite{mescheder2019occupancynet, park2019deepsdf, chen2018implicit_decoder,xie2022neural}, neural fields have become one of the most promising methods for many 3D vision tasks including 3D/4D reconstruction~\cite{yariv2023baked,li2023neuralangelo,park2021hypernerf,niemeyer2019occupancy,park2021nerfies,peng2020convolutional}, 3D generative modeling~\cite{schwarz2020graf,niemeyer2021giraffe,poole2022dreamfusion,chan2022eg3d,Lin2023magic3d,tsalicoglou2023textmesh}, and view synthesis~\cite{mildenhall2020nerf,barron2022mip,Barron2023zipnerf}.
Key to their success is among others simplicity, state-of-the-art performance, and robust optimization~\cite{xie2022neural,wang2024bilateral,brualla2021wild}.
In contrast to previous representations such as point-~\cite{qi2017pointnet,qi2017pointnetpp,peng2021shape}, voxel-~\cite{maturana2015voxnet,qi2016volumetric}, or mesh-based~\cite{wang2018pixel2mesh,groueix2018papier} representations, neural fields do not usually require complex regularization, hand-tuned initialization or optimization control modules as they admit end-to-end optimization and can be queried at arbitrary spatial locations.
In the context of view synthesis, Neural Radiance Fields~\cite{mildenhall2020nerf} (NeRF) in particular have revolutionized the field by leveraging volumetric rendering, which has proven more robust than prior surface-based rendering approaches~\cite{niemeyer2020dvr,yariv2020multiview,sitzmann2019scene}.
In this work, we employ the robustness and simplicity of neural fields to enable real-time rendering for complex scenes at high quality.
More specifically, we use the state-of-the-art radiance field \zipnerf{}~\cite{Barron2023zipnerf} to act as a robust prior and source of reliable supervision to train a point-based representation better suited for real-time rendering.

\boldparagraph{Neural Fields for Real-Time Rendering} 
NeRF-based models lead to state-of-the-art view synthesis but are typically slow to render so that a variety of works are proposed for speeding up training and inference. 
While neural fields were initially built on large, compute-heavy multi-layer perceptrons (MLPs)~\cite{mescheder2019occupancynet, park2019deepsdf, chen2018implicit_decoder,mildenhall2020nerf},
recent works propose the use of voxel representations and interpolation to enable fast training and rendering~\cite{yu2021plenoxels,yu2021plenoctrees,reiser2021kilonerf,mueller2022instant}.
Instant NGP~\cite{mueller2022instant}, for example, demonstrates that a multi-resolution hash grid backbone enables higher quality whilst reducing training time to seconds.
However, these works rely on powerful GPUs and often do not achieve real-time rendering for arbitrary scenes.
Another line of work aims to represent neural fields as meshes, either as a post-processing step~\cite{rakotosaona2023nerfmeshing,tang2022nerf2mesh} or by direct optimization~\cite{yariv2023baked,guo2023vmesh,wang2023shells,turki2024hybridnerf,chen2023mobilenerf,wang2023adaptive,reiser2024binary}.
These approaches can achieve high frame rates but their quality lacks behind volumetric approaches.
More recently, another line of work~\cite{hedman2021baking,reiser2023merf,duckworth2023smerf,garbin2021fastnerf} aims to represent a neural field as a set of easily-cacheable assets such as sparse voxel grids, triplanes, and occupancy grids.
These methods retain their high quality but often exhibit large storage requirements, are slower to render on smaller devices, and rely on complex custom rendering implementations~\cite{duckworth2023smerf}.
In contrast, we optimize lightweight point-based representations that achieve state-of-the-art quality, are easily compressed, and naturally integrate with graphics software following a rasterization pipeline.

\boldparagraph{Point-Based Representations}
First works propose to render point sets as independent geometry samples~\cite{gross2011point,grossman1998point}, which can be implemented efficiently in graphics software~\cite{sainz2004point} and highly parallelized on GPU hardware~\cite{laine2011high,schutz2022software}. %
To eliminate holes when rendering incomplete surfaces, a line of works explores the ``splatting'' of points with extents larger than a pixel, \eg with circular or elliptic shapes~\cite{wiles2020synsin,yifan2019differentiable}.
The recent work 3D Gaussian Splatting (3DGS)~\cite{kerbl20233d} achieves unprecedented quality and fast training and rendering speed by introducing adaptive density control in combination with efficient rasterization kernels.
As a consequence, 3DGS is used in a variety of applications, including 3D human~\cite{zheng2024gpsgaussian} and avatar reconstruction~\cite{dhamo2023headgas, qian2024gaussianavatars, moreau2024human}, 3D generation~\cite{EnVision2024luciddreamer, tang2024dreamgaussian,yi2024gaussiandreamer,chung2023luciddreamer}, SLAM systems~\cite{yan2023gs, li2024sgs, matsuki2024gaussian,splatsplam2024sand}, 4D reconstruction~\cite{wu20244dgaussians,xu2024fourkfourd}, and open-set segmentation~\cite{shi2024language,qin2024langsplat}.
Further, works are proposed to address  aliasing~\cite{Yu2023MipSplatting, yan2023multi} and point densification~\cite{cheng2024gaussianpro,Franke2023vet} in the 3DGS representation.
Finally, a recent line of works investigate compression for 3DGS~\cite{fan2023lightg,lee2023compactgs,morgenstern2023compact,niedermayr2023compressed,navaneet2023compact3d,fang2024minisplatting} and the concurrent work \cite{Foroutan2024arxiv} analyzes initialization alternatives. 
In this work, we combine a NeRF prior for stable optimization with a point-based 3DGS representation for real-time rendering of complex scenes.
Compared to prior works, we enable high-quality view synthesis even for complex real-world captures that might contain lighting and exposure variations.
Further, we develop pruning and test-time visibility rendering strategies leading to 10$\times$ fewer Gaussian primitives at higher quality compared to 3DGS and with inference times of 900+ FPS.

\section{Method}
\label{sec:method}

\begin{figure*}
    \centering
    \includegraphics[width=1.\linewidth]{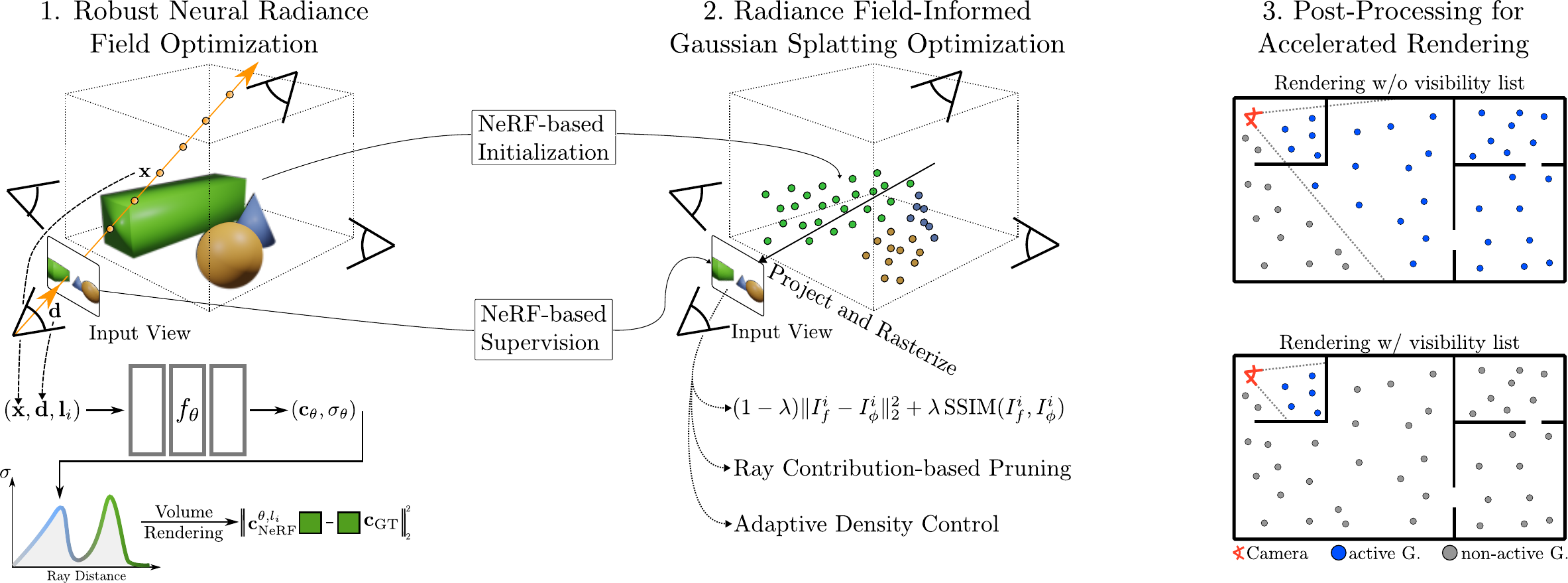}
    \caption{
    \textbf{Overview.} 1.\ Given posed input images of a scene, we train a robust neural radiance field with GLO embeddings $\mathbf{l}_i$.
    2.\ We use the radiance field prior to initialize and supervise our point-based 3DGS representation that we optimize with a novel pruning technique for more compact, high-quality scenes.
    3.\ We perform viewpoint-based visibility filtering to further accelerate test-time rendering speed.  
    }
    \label{fig:overview}
\end{figure*}

Our goal is to develop a lightweight, real-time view synthesis method that is robust even for complex real-world captures. In the following, we discuss the key components for achieving this. 
First, we optimize radiance fields as a robust prior for complex data (\secref{subsec:nerf-prior}).
Next, we use the radiance field to first initialize and then to supervise the optimization of point-based 3DGS representations (\secref{subsec:radsplat}.) 
We develop a novel pruning technique leading to a significant point count reduction while maintaining high quality (\secref{subsec:ray-based-pruning}). 
Finally, we cluster input cameras and perform visibility filtering, further accelerating rendering speed to up to 900+ FPS (\secref{subsec:viewpoint-baking}).
We show an overview of our method in~\figref{fig:overview}.

\subsection{Neural Radiance Fields as a Robust Prior}
\label{subsec:nerf-prior}

\boldparagraph{Neural Radiance Fields} A radiance field $f$ is a continuous function that maps a 3D point $\mathbf{x} \in \mathbb{R}^3$ and a viewing direction $\mathbf{d} \in \mathbb{S}^2$ to a volume density $\sigma \in \mathbb{R}^+$ and an RGB color value $\mathbf{c} \in \mathbb{R}^3$. Inspired by classical volume rendering~\cite{Kajiya1984ray}, a pixel's final color prediction is obtained by approximating the integral via quadrature using sample points:
\begin{align}
    \begin{split}\label{eq:render-nerf}
        &\bc_{\text{NeRF}} = \sum_{j=1}^{N_s} \tau_j \alpha_j \bc_j \quad \text{with} \\
        &\tau_j = \prod_{k=1}^{j-1} (1 - \alpha_k), \quad \alpha_j = 1 - e^{-\sigma_j\delta_j}
    \end{split}
\end{align}
where $\tau_j$ is the transmittance, $\alpha_j$ the alpha value for $\mathbf{x}_j$, and $\delta_j = {\left|\left|\mathbf{x}_{j+1} - \mathbf{x}_j \right|\right|}_2$ the distance between neighboring sample points.
In Neural Radiance Fields~\cite{mildenhall2020nerf}, $f$ is parameterized as an MLP with ReLU activation $f_\theta$ and the network parameters $\theta$ are optimized using gradient descent on the reconstruction loss:
\begin{align}
\mathcal{L} (\theta) = \sum_{\br \in \cR_\text{batch}} \lVert \bc_\text{NeRF}^{\theta}(\br) - \bc_\text{GT}(\br) \rVert_{2}^{2}
\end{align}
where $\br \in \cR_\text{batch}$ are batches of rays sampled from the set of all pixels / rays $\cR$. To further boost training time and quality, \zipnerf{}~\cite{Barron2023zipnerf} uses multisampling and an efficient multi-resolution grid backbone~\cite{mueller2022instant}.
Due to the state-of-the-art performance, we adopt \zipnerf{} as our radiance field prior (see supp.\ mat.\ for a comparison to an INGP~\cite{mueller2022instant} backbone).

\boldparagraph{Robust Optimization on Real-World Data} Real-world captures often contain effects such as lighting and exposure variation or motion blur. 
Crucial for the success of neural fields on such in-the-wild data~\cite{brualla2021wild} is the use of Generative Latent Optimization~\cite{Bojanowski2018glo} (GLO) embedding vectors or related techniques.
More specifically, a per-image latent vector is optimized along with the neural field that enables explaining away these image- and view-dependent effects
\begin{align}
    \cL(\theta, \{ \mathbf{l}_i\}_{i=1}^{N} ) = \sum_{\br_i \in \cR_\text{batch}} \lVert \bc_\text{NeRF}^{\theta, l_i}(\br_i) - \bc_\text{GT}(\br_i) \rVert_{2}^{2}
\end{align}
where $\{ \mathbf{l}_i\}_{i=1}^{N}$ indicates the set of GLO vectors and $N$ the number of input images.
This allows the model to express appearance changes captured in the input images without introducing wrong geometry such as floating artifacts.
At test time, images can be rendered with a constant latent vector (usually the zero vector) to obtain stable and high-quality view synthesis.
For all experiments, we follow~\cite{Barron2023zipnerf} and optimize a per-image latent vector representing an affine transformation for the bottleneck vector in the \zipnerf{} representation.

\subsection{Radiance Field-Informed Gaussian Splatting}
\label{subsec:radsplat}

\boldparagraph{Gaussian Splatting} In contrast to neural fields, in 3D Gaussian Splatting~\cite{kerbl20233d} an explicit point-based scene representation is optimized. More specifically, the scene is represented as points that are associated with a position $\bp \in \nR^3$, opacity $o \in [0, 1]$, third-degree spherical harmonics (SH) coefficients $\bk \in \nR^{16}$, 3D scale $\bs \in \nR^3$, and the 3D rotation $R \in SO(3)$ represented by 4D quaternions $\bq \in \nR^{4}$.
Similar to~\eqref{eq:render-nerf}, such a representation can be rendered to the image plane for a camera and a list of correctly-sorted points as
\begin{align}\label{eq:render-gs}
    \bc_\text{GS} = \sum_{j=1}^{N_p} \bc_j \alpha_j \tau_i \quad \text{where}\quad \tau_i=\prod_{i=1}^{j-1} (1 - \alpha_i)
\end{align}
where $\bc_j$ is the color predicted using the SH coefficients $\bk$ and $\alpha_j$ is obtained by evaluating the projected 2D Gaussian with covariance $\Sigma' = JM \Sigma M^TJ^T$, multiplied by the per-point opacity $o$~\cite{kerbl20233d}, with $M$ being the viewing transformation, $J$ denoting the Jacobian of the affine approximation of the projective transformation~\cite{zwicker2001ewa}, and $\Sigma$ denoting the 3D covariance matrix. To ensure that $\Sigma$ is a positive semi-definite matrix, it is expressed using the per-point scale matrix $S = \text{diag}(s_1, s_2, s_3)$ and rotation $R$ according to $\Sigma = RSS^TR^T$~\cite{kerbl20233d}. The scene is optimized with a reconstruction loss on the input images and regular densification steps consisting of splitting, merging, and pruning points based on gradient and opacity values. 

\boldparagraph{Radiance Field-based Initialization}
A key strength of radiance fields lies in the volume rendering paradigm~\cite{mildenhall2020nerf}, as opposed to prior surface rendering techniques~\cite{niemeyer2020dvr,yariv2020multiview,sitzmann2019scene}, enabling the ability to initialize, remove, and change density freely in 3D space. 
In contrast, explicit point-based representations can only provide a gradient signal to already existing geometry prediction due to the rasterization-based approach.
The initialization of this representation is hence a crucial property in its optimization process.

We propose to use the radiance field prior for obtaining a suitable initialization. More specifically, for each pixel / ray $\br$ we first define the median depth $z_\text{median}$ of our NeRF model as the distance 
to the first sample along the ray with
accumulated transmittance $\tau_i > 0.5$.
We unproject all pixels / rays into 3D space to obtain our initial point set
\begin{align}
  \cP_\text{init} = \{\bp_i\}_{i \in \cK_\text{rnd}}, \, \bp_i = \br_0(i) + \bd_{r(i)} \cdot z_\text{median}(\br(i))
\end{align}
where $\cK_\text{rnd}$ are  uniform randomly-sampled indices for the list of all rays / pixels, $\br_0(\cdot)$ indicates the ray origin and $\bd_{r(\cdot)}$ the normalized ray direction. We found the median depth estimation to perform better than other common choices such as expected depth by being exact sampling point estimates, and we found setting $\vert\cK_\text{rnd}\vert$ to $1$ million for all scenes to work well in practice (see Sec.\ 3.2 of supp.\ mat.\ for more details).
Further, we initialize
\begin{align}
\begin{split}
    \bk_i &= (\bk_i^{1:3}, \bk_i^{4:16}) \,,\, \bk_i^{1:3} = \bc_\text{NeRF}(\br(i)),\, \bk_i^{4:16} = \mathbf{0}\footnotemark \\
    \bs_i &= (s_i, s_i, s_i) \, , \, s_i = \min_{p \in \left\{p \neq p_i \vert p \in \cP_\text{init} \right\} } \lVert \bp_i - \bp \rVert_2
\end{split}
\end{align}\footnotetext{We found progressively optimizing $\bk^{4:16}$ leads to better results~\cite{kerbl20233d}.}
and set $o_i = 0.1$ and $\bq_i$ to the identity rotation. Thus, for each scene we optimize
\begin{align}
    \phi = \{(\bp_i, \bk_i, \bs_i, o_i, \bq_i)\}_{i=1}^{N_\text{init}}
\end{align}

\boldparagraph{Radiance Field-based Supervision} Radiance fields have been shown to excel even on real-world captures where images contain challenging exposure and lighting variations~\cite{brualla2021wild,Barron2023zipnerf}. We leverage this strength of radiance fields to factor out this complexity and noise of the data to provide a more cleaned up supervision signal than the possibly corrupted input images. More specifically, we render all input images with our NeRF model $f_\theta$ and with a constant zero GLO vector
\begin{align}
\cI_{f} = \{ I_f^j \}_{j=1}^{N} \quad \text{where} \quad
    I_f^j = \{ \bc_\text{NeRF}^{\theta,l_\text{zero}}({\br_j(i)}) \}_{i=1}^{H\times W}
\end{align}
where $\mathbf{l}_\text{zero}$ indicates the zero GLO vector, $H$ the height and $W$ the width of the images, and $\br_j(\cdot)$ the rays belonging to the $j$-th image.
We can then use these renderings $\cI_f$ to train our point-based representations
\begin{align}
    \begin{split}
        \cL(\phi) = (1 - \lambda) \lVert I_f^i - I_\phi^i \rVert_2^2 + \lambda \, \text{SSIM}(I_f^i, I_\phi^i)
    \end{split}
\end{align}
where $i \sim \cU$ is drawn from the uniform distribution and we use the default value $\lambda = 0.2$.
Another practical benefit of this approach is that we can train from arbitrary camera lens types due to NeRF's flexible ray casting, while the 3D Gaussian Splatting gradient formulation assumes a pinhole camera model and it is unclear how this can be efficiently extended to \eg fisheye or more complex lens types.

\subsection{Ray Contribution-Based Pruning}
\label{subsec:ray-based-pruning}

While 3DGS representations can be efficiently rendered thanks to rasterization, real-time performance still requires a powerful GPU and is not yet achieved on all platforms.
The most important property for the rendering performance is the number of points in the scene that need to be rendered. 

\boldparagraph{Importance Score} To obtain a more lightweight representation that can be rendered faster across platforms, we develop a novel pruning technique to reduce the number of Gaussians in the scene whilst maintaining high quality.
More specifically, we introduce a pruning step during optimization that removes points that do not contribute significantly to any training view.
To this end, we define an importance score by aggregating the ray contribution of Gaussian $\bp_i$ along all rays of all input images
\begin{align}\label{eq:importance-score}
    h(\bp_i) = \max_{I_f \in \cI_f, r \in I_f} \alpha_i^r \tau_i^r
\end{align}
where $\alpha_i^r \tau_i^r$ indicates the ray contribution for the pixel's final color prediction in~\eqref{eq:render-gs} of Gaussian $\bp_i$ along ray $\br$.
We find that this formulation leads to improved results compared to concurrent works that investigate similar ideas~\cite{lee2023compactgs,fan2023lightg} as we use the exact ray contribution (as opposed to \eg the opacity) as well as the max operator (instead of \eg the mean) which is independent of the number of input images, hence more robust to different types of scene coverage~\cite{Goli2023nerf2nerf}.

\boldparagraph{Pruning} We use our importance score during optimization to reduce the overall point count in the scene while maintaining high quality. More specifically, we add a pruning step where we calculate mask values as
\begin{align}
    m_i = m(\bp_i) = \mathbb{1} \left( h(\bp_i) < t_\text{prune} \right),\, t_\text{prune} \in [0, 1]
\end{align}
and we remove all Gaussians from our scene that have a mask value of one.
We apply the pruning step twice over the the course of optimization similar to~\cite{fan2023lightg}.
The threshold $t_\text{prune}$ provides a control mechanism over the number of points that are used to represent the scene. 
In our experiments, we define two values, one value for our default model, and a higher value for a lightweight variant.

\subsection{Viewpoint-Based Visibility Filtering}
\label{subsec:viewpoint-baking}

Our pruning technique ensures a compact scene representation with a small overall point count.
To scale to larger, more complex scenes such as entire houses or apartments, inspired by classical occlusion culling~\cite{Coorg1997occlusion}, we introduce a novel viewpoint-based filtering as post-processing step that further speeds up test-time rendering without a quality drop. 

\boldparagraph{Input Camera Clustering} First, we group input cameras together to obtain a meaningful tessellation of the scene space. More specifically, let $(\bx_\text{cam}^i)_{i=1}^{N}$ denote the input camera locations for the set of input images $\cI$.
We run $k$-means clustering on the input camera locations to obtain $k$ cluster centers $(\bx_\text{cluster}^i)_{i=1}^{k}$ and assign the input cameras to the respective cluster centers.

\boldparagraph{Visibility Filtering} Next, for each cluster center $\bx_\text{cluster}^j$, we select all assigned input cameras, render the images from these viewpoints, and, similar to~\eqref{eq:importance-score}, calculate an importance score and the respective visibility indicator list
\begin{align}\begin{split}\label{eq:importance-score-vis}
    h^\text{cluster}_j(\bp_i) &= \max_{I \in \cI_c^i, r \in I} \alpha_i^r \tau_i^r, \\
    m^\text{cluster}_j(\bp_i) &= \mathbb{1} \left( h^\text{cluster}_j(\bp_i) > t_\text{cluster} \right)
\end{split}\end{align}
where $\cI_c^i$ is the set of images whose camera positions are assigned to the the cluster center $\bx_\text{cluster}^i$ and $t_\text{cluster}$ is a threshold that controls the contribution value of points that should be filtered out (we found setting $t_\text{cluster} = 0.001$ to work well in practice).
Note that we are not restricted to the input views for calculating these masks. In practice, we hence add random camera samples to $\cI_c^i$ to ensure robustness to test views.
We calculate the indicator list $m_j^\text{cluster}$ for each cluster center as a post-processing step after scene optimization.

\boldparagraph{Visibility List-Based Rendering}
To render an arbitrary viewpoint, we first assign its camera center $\bx_\text{cam}^\text{test}$ to the nearest cluster center $\bx_\text{cluster}^i$. Next, we select the respective indicator list $m_i^\text{cluster}$. Finally, we perform default rasterization while only considering the points that are marked as active for the respective cluster.
This results in a significant FPS increase up to 45\% without any drop in quality.

\subsection{Implementation Details} We set the number of initial points $N_\text{init}$ to $1$ million in all experiments. For threshold value $t_\text{prune}$, we use $0.01$ and $0.25$ for our default and lightweight models, respectively, and for the large \zipnerf{} scenes, we use $0.005$ and $0.03$.
We perform pruning after $16$ and $24$ thousand steps.
We follow~\cite{kerbl20233d} and use the same densification parameters except for the densification gradient threshold value which we lower to $8.6e^{-5}$ for the \zipnerf{} dataset. 
We train our radiance fields on 8 V100 GPUs ($\sim$1h) and our 3DGS models on a A100 GPU ($\sim$1h).
For the visibility filtering, we use $k=64$ clusters and we found a small threshold $t_\text{cluster} > 0$ to work well in practice and set it to $0.001$ for all scenes.
For the radiance field training, we follow~\cite{Barron2023zipnerf} and use default parameters for all scenes.

\section{Experiments}
\label{sec:experiments}

\boldparagraph{Datasets} We report results on the MipNeRF360 dataset~\cite{barron2022mip}, the most common view synthesis benchmark consisting of $9$ unbounded indoor and outdoor scenes.
We further report results on the \zipnerf{} dataset~\cite{Barron2023zipnerf} consisting of $4$ large-scale scenes (apartments and houses) with challenging captures that partly contain lighting and exposure variations.

\boldparagraph{Baselines} On all datasets, we compare against 3DGS~\cite{kerbl20233d} as well as MERF~\cite{reiser2023merf} and SMERF~\cite{duckworth2023smerf} as the state-of-the-art volumetric approaches that construct efficient voxel and triplane representations together with accelerating structures for empty space skipping. %
On MipNeRF360, we further compare against mesh-based BakedSDF~\cite{yariv2023baked}, hash-grid based INGP~\cite{mueller2022instant}, and point-based approaches LightGaussian~\cite{fan2023lightg}, CompactGaussian~\cite{lee2023compactgs}, and EAGLES~\cite{girish2023eagles}. For reference, we always report \zipnerf{}~\cite{Barron2023zipnerf}, the state-of-the-art offline view synthesis method.

\boldparagraph{Metrics and Evaluation} We follow common practice and report the view synthesis metrics PSNR, SSIM, and LPIPS.
While using techniques such as GLO vectors is essential for high quality on real-world captures (see~\secref{subsec:nerf-prior}), the evaluation of such models is an open problem such that recent methods~\cite{duckworth2023smerf,Barron2023zipnerf,barron2022mip} train two separate models, one for visualizations, and one (without GLO vectors) purely for the quantitative comparison.
In this work, we always train a single model that is robust thanks to the radiance field prior. For evaluation, we simply finetune the trained models on the original training set image data to match potential color shifts and to ensure a fair comparison.
Next to measuring quality, we report the rendering speed in frames per second (FPS) on a RTX 3090 GPU and the number of Gaussians in the scenes (only applicable for point-based methods).

\subsection{Real-Time View Synthesis}
\begin{figure*}
\centering
\captionsetup[subfigure]{labelformat=empty}
\begin{subfigure}{0.255\linewidth}
\centering
\includegraphics[width=0.85\linewidth]{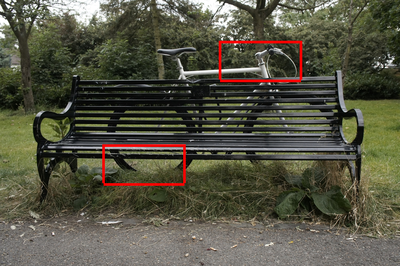}
\subcaption{Bicycle}
\end{subfigure}
\begin{subfigure}{0.177\linewidth}
\includegraphics[width=1.0\linewidth]{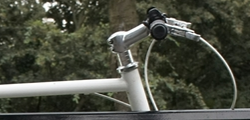}
\includegraphics[width=1.0\linewidth]{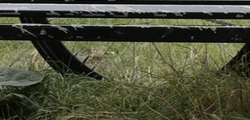}
\end{subfigure}
\begin{subfigure}{0.177\linewidth}
\includegraphics[width=\linewidth]{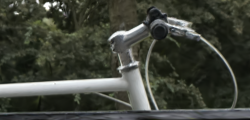}
\includegraphics[width=\linewidth]{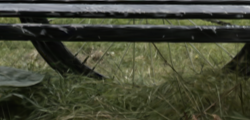}
\end{subfigure}
\begin{subfigure}{0.177\linewidth}    \includegraphics[width=\linewidth]{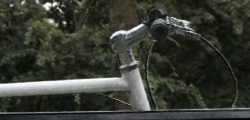}
\includegraphics[width=\linewidth]{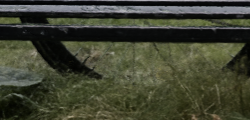}
\end{subfigure}
\begin{subfigure}{0.177\linewidth}    \includegraphics[width=\linewidth]{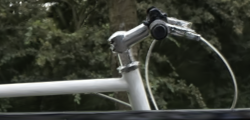}
\includegraphics[width=\linewidth]{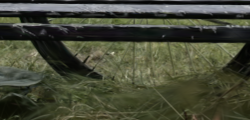}
\end{subfigure}
\begin{subfigure}{0.255\linewidth}
\centering
\includegraphics[width=0.85\linewidth]{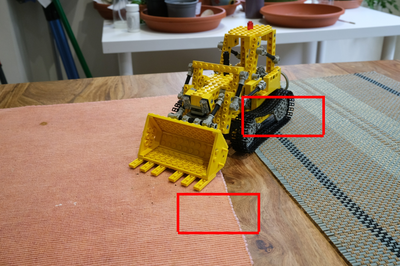}
\subcaption{Kitchen}
\end{subfigure}
\begin{subfigure}{0.177\linewidth}
\includegraphics[width=1.0\linewidth]{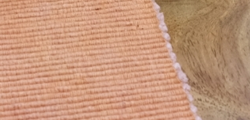}
\includegraphics[width=1.0\linewidth]{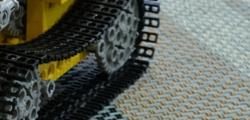}
\end{subfigure}
\begin{subfigure}{0.177\linewidth}
\includegraphics[width=\linewidth]{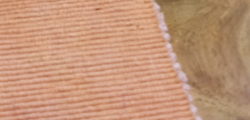}
\includegraphics[width=\linewidth]{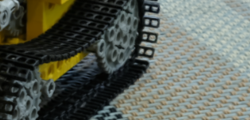}
\end{subfigure}
\begin{subfigure}{0.177\linewidth}    \includegraphics[width=\linewidth]{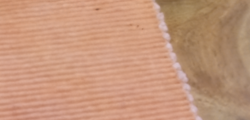}
\includegraphics[width=\linewidth]{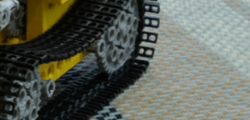}
\end{subfigure}
\begin{subfigure}{0.177\linewidth}    \includegraphics[width=\linewidth]{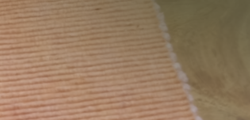}
\includegraphics[width=\linewidth]{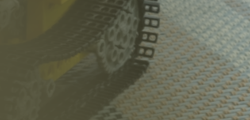}
\end{subfigure}
\begin{subfigure}{0.255\linewidth}
\centering
\includegraphics[width=0.85\linewidth,trim={1.2cm 0 1.2cm 0},clip]{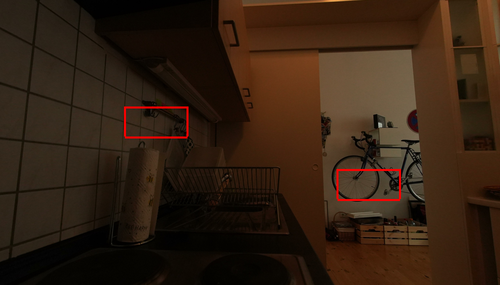}
\subcaption{Berlin}
\end{subfigure}
\begin{subfigure}{0.177\linewidth}
\includegraphics[width=1.0\linewidth]{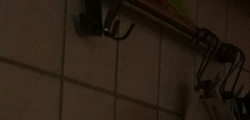}
\includegraphics[width=1.0\linewidth]{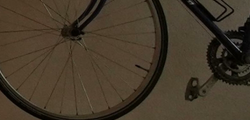}
\end{subfigure}
\begin{subfigure}{0.177\linewidth}
\includegraphics[width=\linewidth]{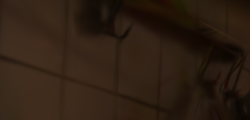}
\includegraphics[width=\linewidth]{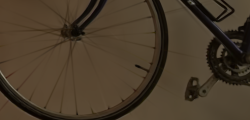}
\end{subfigure}
\begin{subfigure}{0.177\linewidth}    \includegraphics[width=\linewidth]{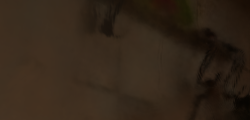}
\includegraphics[width=\linewidth]{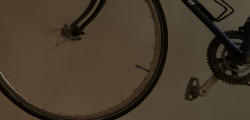}
\end{subfigure}
\begin{subfigure}{0.177\linewidth}    \includegraphics[width=\linewidth]{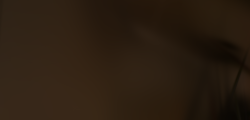}
\includegraphics[width=\linewidth]{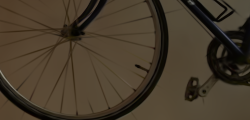}
\end{subfigure}
\begin{subfigure}{0.255\linewidth}
\centering
\includegraphics[width=0.85\linewidth,trim={1.2cm 0 1.2cm 0},clip]{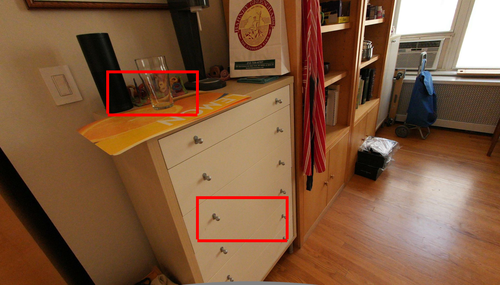}
\subcaption{NYC}
\end{subfigure}
\begin{subfigure}{0.177\linewidth}
\includegraphics[width=1.0\linewidth]{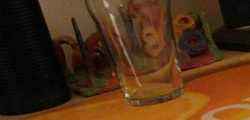}
\includegraphics[width=1.0\linewidth]{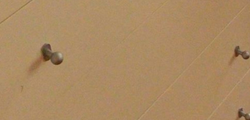}
\end{subfigure}
\begin{subfigure}{0.177\linewidth}
\includegraphics[width=\linewidth]{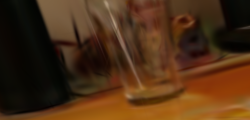}
\includegraphics[width=\linewidth]{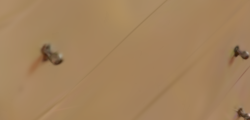}
\end{subfigure}
\begin{subfigure}{0.177\linewidth}    \includegraphics[width=\linewidth]{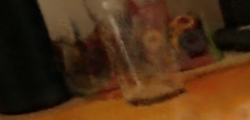}
\includegraphics[width=\linewidth]{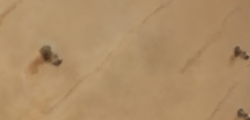}
\end{subfigure}
\begin{subfigure}{0.177\linewidth}    \includegraphics[width=\linewidth]{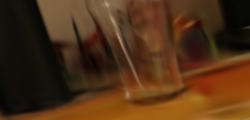}
\includegraphics[width=\linewidth]{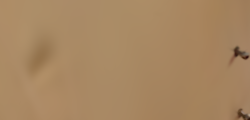}
\end{subfigure}
\begin{subfigure}{0.255\linewidth}
\centering
\includegraphics[width=0.85\linewidth,trim={1.2cm 0 1.2cm 0},clip]{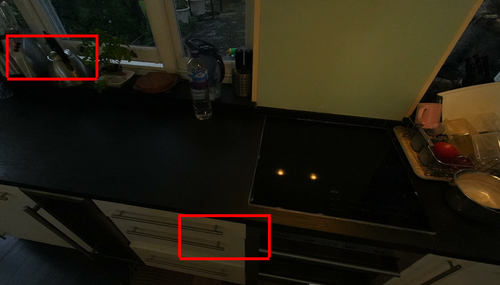}
\subcaption{London}
\end{subfigure}
\begin{subfigure}{0.177\linewidth}
\includegraphics[width=1.0\linewidth]{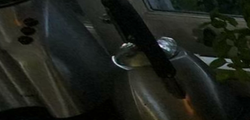}
\includegraphics[width=1.0\linewidth]{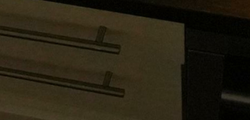}
\subcaption{GT}
\end{subfigure}
\begin{subfigure}{0.177\linewidth}
\includegraphics[width=\linewidth]{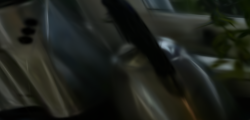}
\includegraphics[width=\linewidth]{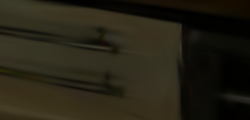}
\subcaption{\textbf{Ours}}
\end{subfigure}
\begin{subfigure}{0.177\linewidth}    \includegraphics[width=\linewidth]{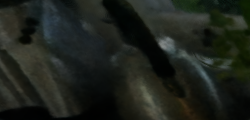}
\includegraphics[width=\linewidth]{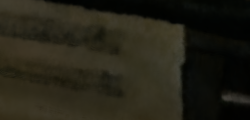}
\subcaption{\zipnerf{}~\cite{Barron2023zipnerf}}
\end{subfigure}
\begin{subfigure}{0.177\linewidth}    \includegraphics[width=\linewidth]{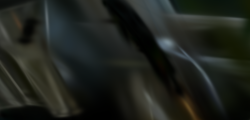}
\includegraphics[width=\linewidth]{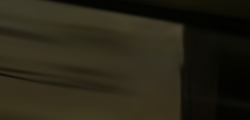}
\subcaption{3DGS~\cite{kerbl20233d}}
\end{subfigure}
\vspace{-.25cm}
\caption{
    \textbf{Qualitative Comparison.}
    We show results on Bicycle and Kitchen from~\cite{barron2022mip} and on Berlin, NYC, London from~\cite{Barron2023zipnerf}. %
    Compared to \zipnerf{}, our method better captures high-frequency textures (\eg, tablecloth in Kitchen) and geometric details (\eg, bicycle spokes in Berlin). Compared to 3DGS, we obtain sharper (\eg, shiny surfaces in London) and more stable results (\eg, color shift in Kitchen).
}
\label{fig:mipnerf360}
\end{figure*}

\begin{table}[ht]
\centering
\begin{subtable}[b]{.97\textwidth}
\resizebox{1.0\textwidth}{!}{
\begin{tabular}{l|cccccc}
\multicolumn{1}{c|}{} & SSIM$\uparrow$  & PSNR$\uparrow$ & LPIPS$\downarrow$ & FPS$\uparrow$ & \#G(M)$\downarrow$
\\\hline
INGP~\cite{mueller2022instant} & 0.705 & 25.68 & 0.302 & 9.26 & - \\
BakedSDF~\cite{yariv2023baked} & 0.697 & 24.51 & 0.309 & \cellcolor{orange!25} 539 & - \\
MERF~\cite{reiser2023merf} & 0.722 & 25.24 & 0.311 & 171 & - \\
SMERF~\cite{duckworth2023smerf} & \cellcolor{yellow!25} 0.818 & \cellcolor{orange!25} 27.99 & \cellcolor{orange!25} 0.211 & 228 & - \\
\hline
CompactG~\cite{lee2023compactgs} & 0.798 & 27.08 & 0.247 & 128 & \cellcolor{yellow!25} 1.388\\ 
LightG~\cite{fan2023lightg} & 0.799 & 26.99 & 0.25 & 209 & \cellcolor{orange!25} 1.046 \\
EAGLES~\cite{girish2023eagles} & 0.809 & 27.16 & 0.238 & 137 & 1.712 \\
3DGS~\cite{kerbl20233d} & 0.815 & 27.20 & 0.214 & 251 & 3.161 \\
\textbf{Ours Light} & \cellcolor{orange!25} 0.826 & \cellcolor{yellow!25} 27.56 & \cellcolor{yellow!25} 0.213 & \cellcolor{red!25} 907 & \cellcolor{red!25} 0.370 \\
\textbf{Ours} & \cellcolor{red!25} 0.843 & \cellcolor{red!25} 28.14 & \cellcolor{red!25} 0.171 & \cellcolor{yellow!25} 410 & 1.924 \\
\hline
\zipnerf{}~\cite{Barron2023zipnerf} & 0.836 & 28.54 & 0.177 & 0.25 & - \\
\hline
\end{tabular}
}
\caption{Mip-NeRF360 dataset~\cite{barron2022mip}}
\label{tab:mipnerf360}
\end{subtable}
\hfill
\begin{subtable}[b]{.9\linewidth}
\resizebox{1.0\textwidth}{!}{
\begin{tabular}{l|cccccc}
\multicolumn{1}{c|}{} & SSIM$\uparrow$  & PSNR$\uparrow$ & LPIPS$\downarrow$ & FPS$\uparrow$
\\\hline
MERF~\cite{reiser2023merf} & 0.747 & 23.49 & 0.445 & 318  \\
SMERF~\cite{duckworth2023smerf} ($K = 1$) & 0.776 & 25.44 & 0.412 & 356 \\
SMERF~\cite{duckworth2023smerf} ($K = 5$) & \cellcolor{yellow!25} 0.829 & \cellcolor{red!25} 27.28 &  \cellcolor{red!25} 0.340 & 221  \\
\hline
3DGS~\cite{kerbl20233d} & 0.809 & 25.50 & 0.369 & \cellcolor{yellow!25} 470\\
\textbf{Ours Light} & \cellcolor{orange!25} 0.838 & \cellcolor{yellow!25} 26.11 & \cellcolor{yellow!25} 0.368 & \cellcolor{red!25} 748 \\
\textbf{Ours} & \cellcolor{red!25} 0.839 & \cellcolor{orange!25} 26.17 & \cellcolor{orange!25} 0.364 & \cellcolor{orange!25} 630 \\
\hline
\zipnerf{}~\cite{Barron2023zipnerf} & 0.836 & 27.37 & 0.305 & 0.25 \\
\hline
\end{tabular}
}
\caption{\zipnerf{} dataset~\cite{Barron2023zipnerf}}
\label{tab:zipnerf}
\end{subtable}
\caption{
\textbf{Quantitative Comparison.}
We compare top-performing real-time rendering approaches and report offline method ZipNeRF as reference.
Our models outperform both NeRF- and GS-based approaches, achieving state-of-the-art view synthesis at higher FPS. %
Ours Light achieves a 10$\times$ reduction of Gaussians (\#G) compared to 3DGS while improving quality (\ref{tab:mipnerf360}).
Our default model improves even over \zipnerf{} in SSIM and LPIPS while rendering 3,600$\times$ faster.
On the large-scale scenes in~\ref{tab:zipnerf}, our models produce the highest SSIM while rendering up to 3.3$\times$ faster than top-performing real-time methods such as SMERF. 
}
\end{table}

\boldparagraph{Unbounded Scenes}
We observe in~\tabref{tab:mipnerf360} that our method leads to the best quantitative results while achieving faster rendering times than prior state-of-the-art real-time methods such as SMERF~\cite{duckworth2023smerf}.
Notably, our model even outperforms the state-of-the-art non-real-time method \zipnerf{}~\cite{Barron2023zipnerf} in both SSIM and LPIPS while rendering 1,600$\times$ faster.
Our lightweight variant (``Ours Light'') also exceeds prior works with a mean rendering speed of 907 FPS outpacing even state-of-the-art mesh-based methods such as BakedSDF~\cite{yariv2023baked}.
Also qualitatively in~\figref{fig:mipnerf360}, we observe that our model achieves the best results.
Compared to \zipnerf{}, our method better captures high-frequency textures (\eg, see tablecloth in ``Kitchen'' scene in~~\figref{fig:mipnerf360}) and fine geometric details (\eg, see bicycle spokes in ``Bicycle'' scene in~\figref{fig:mipnerf360}).
Compared to 3DGS, we find that our reconstructions are sharper and more stable %
while achieving a 2$\times$ and 10$\times$ overall point count reduction with our default and lightweight variant, respectively.

\boldparagraph{Large-Scale Scenes}
For the \zipnerf{} dataset~\cite{Barron2023zipnerf}, we observe a similar trend in~\tabref{tab:zipnerf}. Our default and lightweight variant outperform top-performing real-time SMERF and non-real-time \zipnerf{} in SSIM while rendering significantly faster. 
Notably, our lightweight variant achieves high quality with a mean SSIM of $0.838$ while rendering on average at 748 FPS.
In contrast, the state-of-the-art real time method for large scenes, \ie the large variant of SMERF~\cite{duckworth2023smerf} with $5^3=125$ submodels, achieves a slightly lower SSIM of $0.829$ with a rendering speed of 221 FPS. 
Also qualitatively in~\figref{fig:mipnerf360}, we observe that our model achieves high visual appeal with sharper and more stable reconstructions. %
In contrast to 3DGS~\cite{kerbl20233d}, we find that our method is more robust on challenging captures as shown in~\figref{fig:robustness} where 3DGS leads to heavily degraded results on the Alameda scene.
Note that for 3DGS, results still contain floating artifacts, even when equipped with a per-image module that can handle exposure and lighting variations~\cite{kerbl20233d,duckworth2023smerf}. In contrast, our method enables high-quality synthesis even for in-the-wild data.

\subsection{Ablation Study and Limitations}

\begin{figure}[ht]
\centering
\begin{subfigure}{1.0\textwidth}
\begin{subfigure}{0.32\textwidth}
\includegraphics[width=1.0\linewidth]{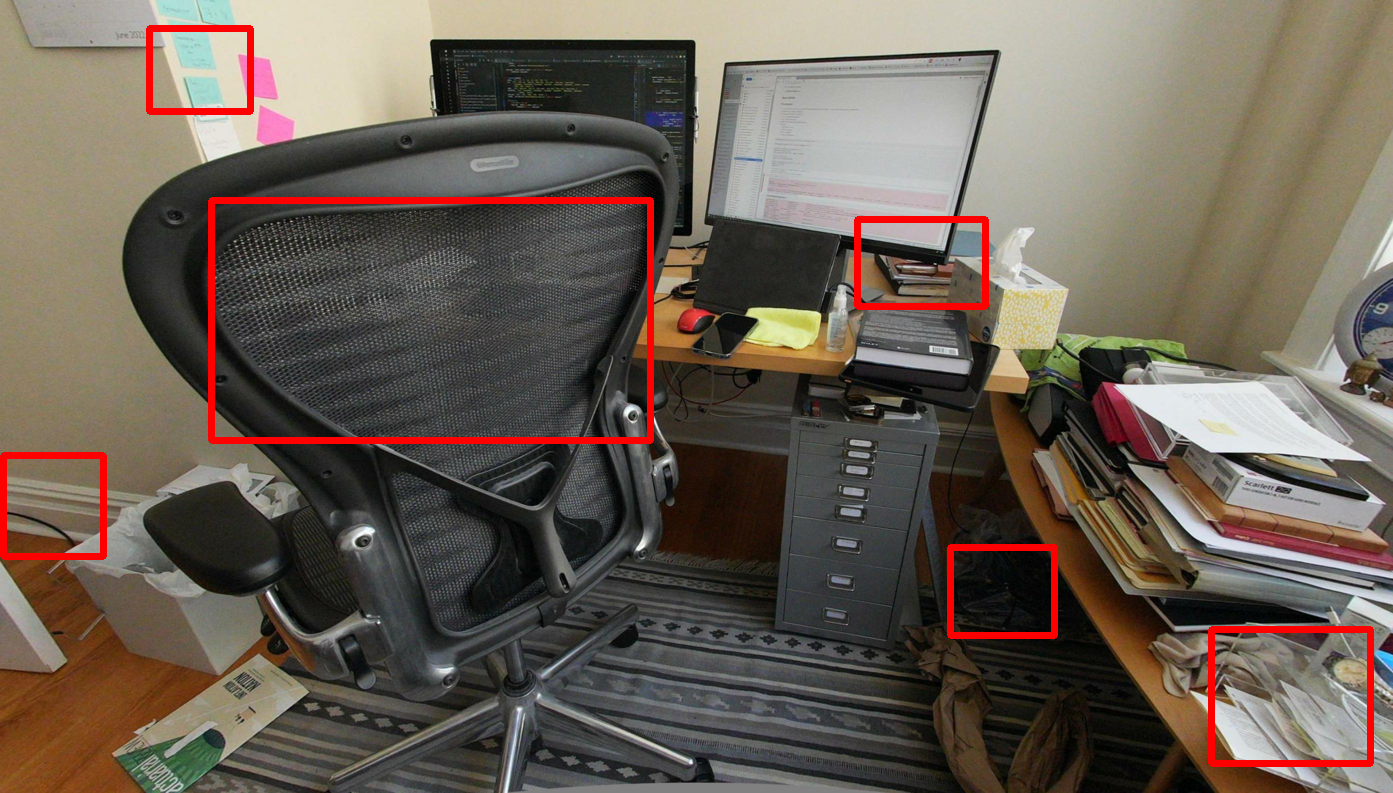}
\subcaption*{GT}
\end{subfigure}
\begin{subfigure}{0.32\textwidth}
\includegraphics[width=\linewidth]{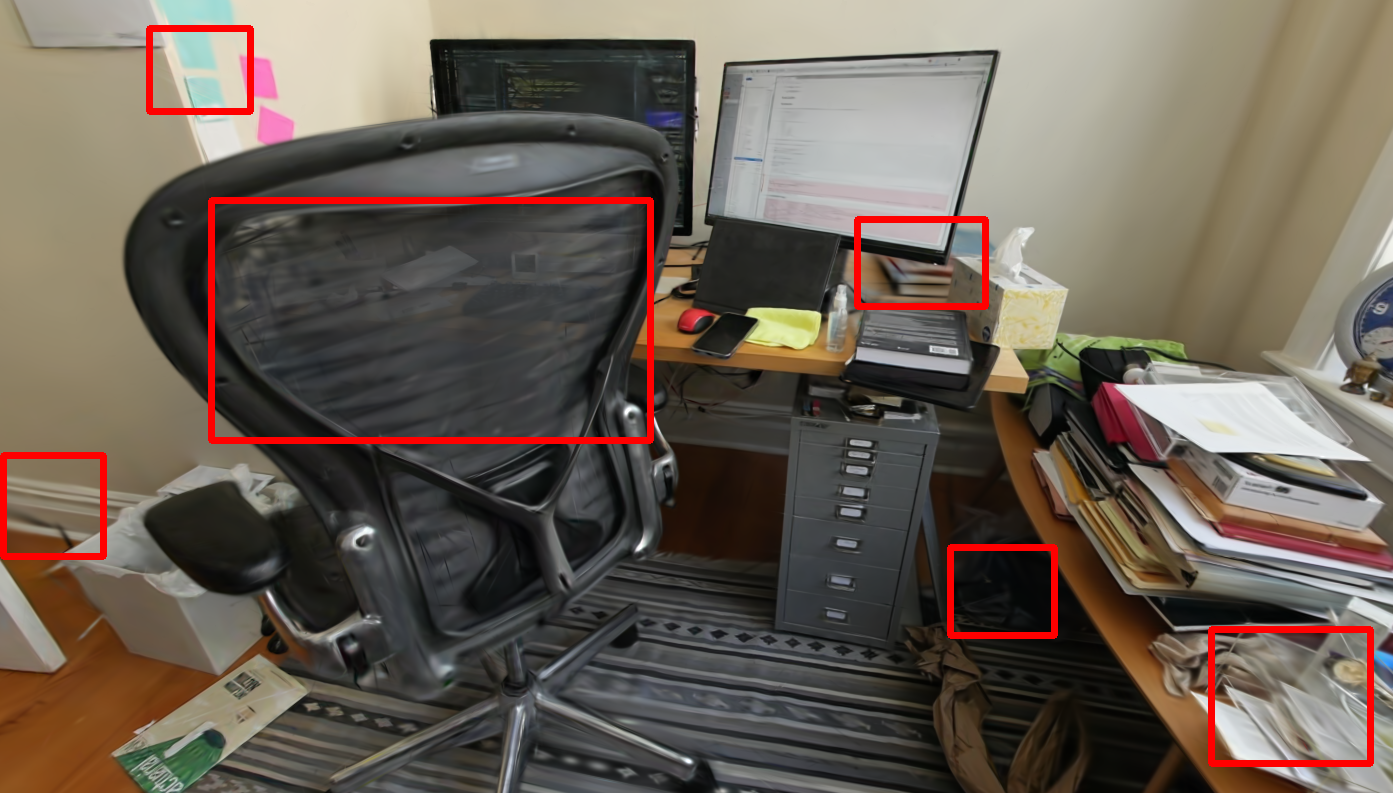}
\subcaption*{\textbf{Ours}}
\end{subfigure}
\begin{subfigure}{0.32\textwidth}
\includegraphics[width=\linewidth]{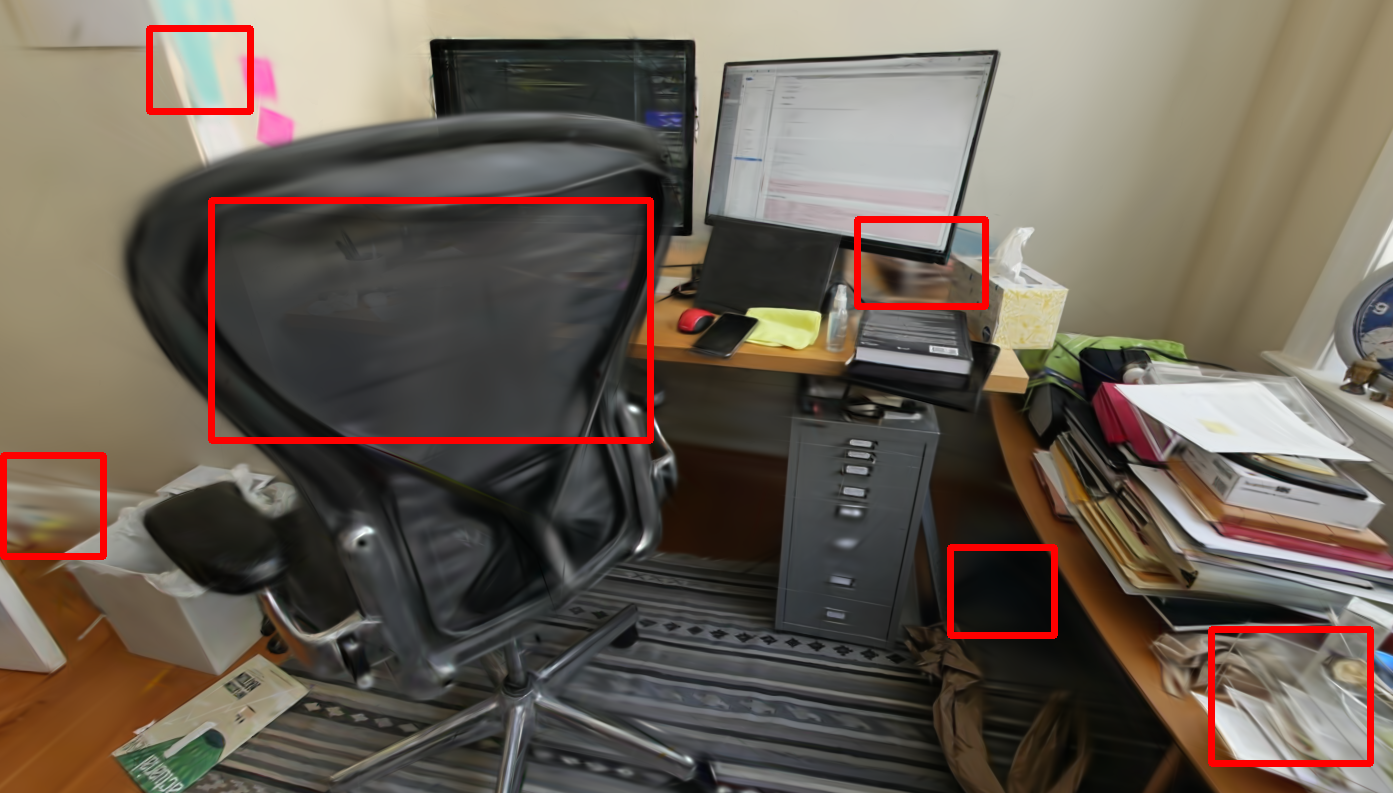}
\subcaption*{w/o NeRF Initialization}
\end{subfigure}
\subcaption{Qualitative Ablation Study of the NeRF-based Initialization.}\label{subfig:ablation1}
\end{subfigure}
\begin{subfigure}{1.0\linewidth}
\begin{subfigure}{0.32\textwidth}
\includegraphics[width=\linewidth]{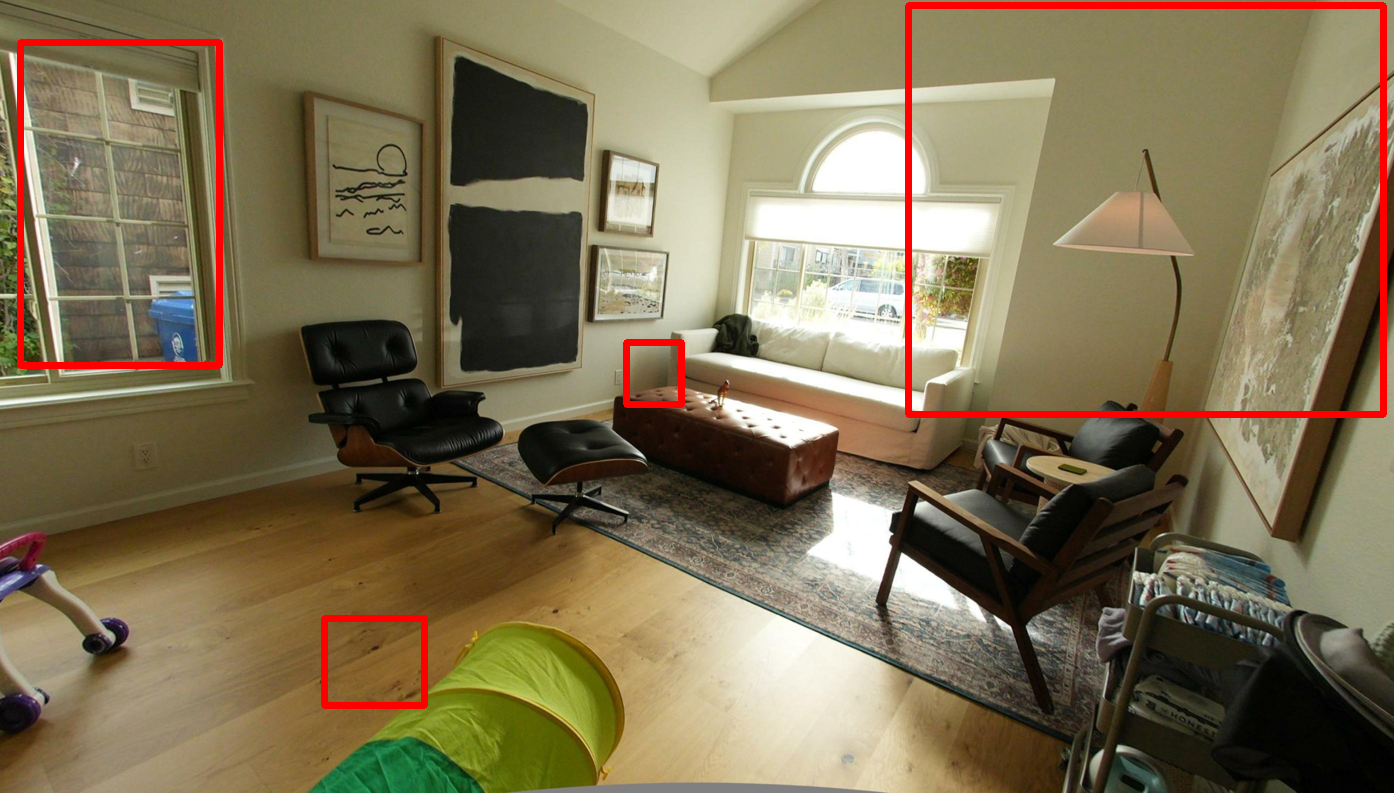}
\subcaption*{GT}
\end{subfigure}
\begin{subfigure}{0.32\textwidth}
\includegraphics[width=\linewidth]{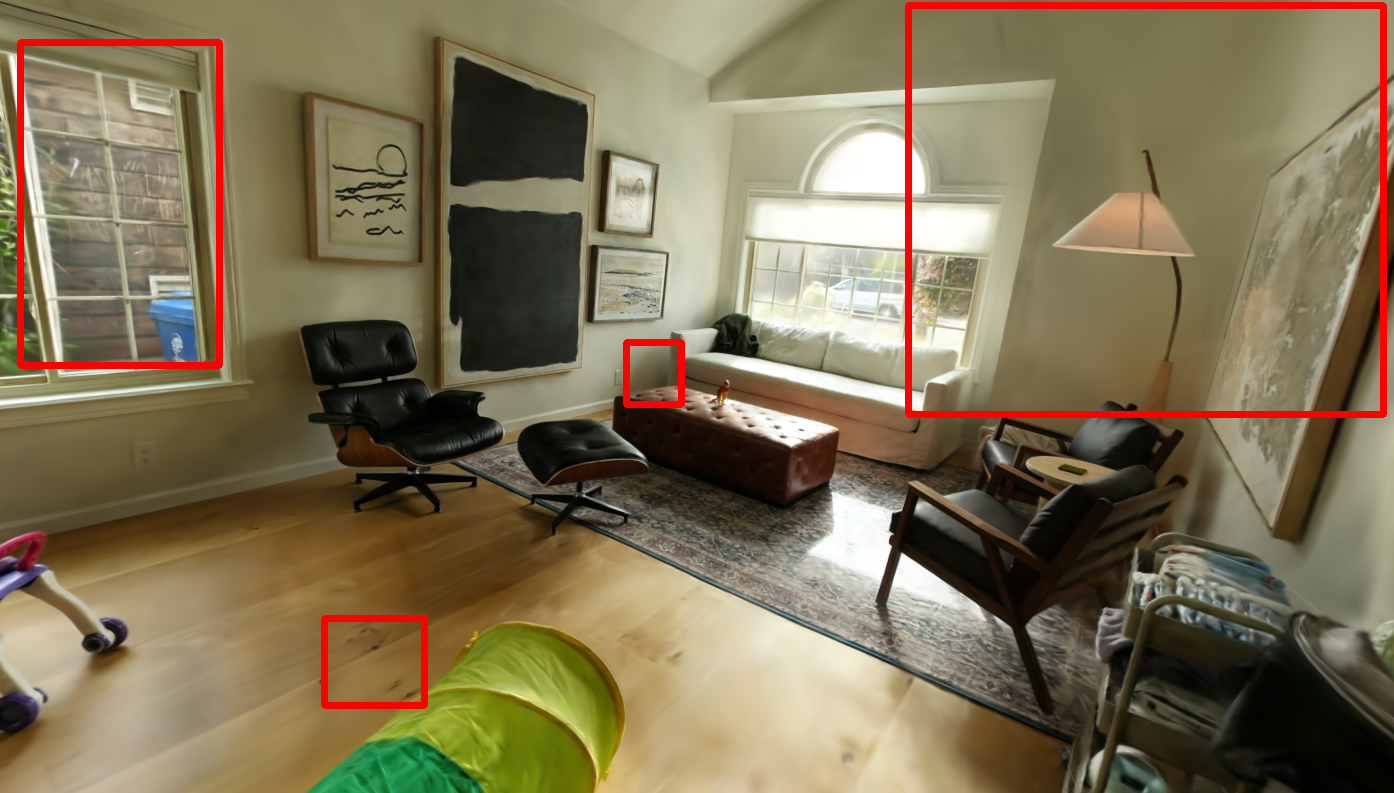}
\subcaption*{\textbf{Ours}}
\end{subfigure}
\begin{subfigure}{0.32\textwidth}
\includegraphics[width=\linewidth]{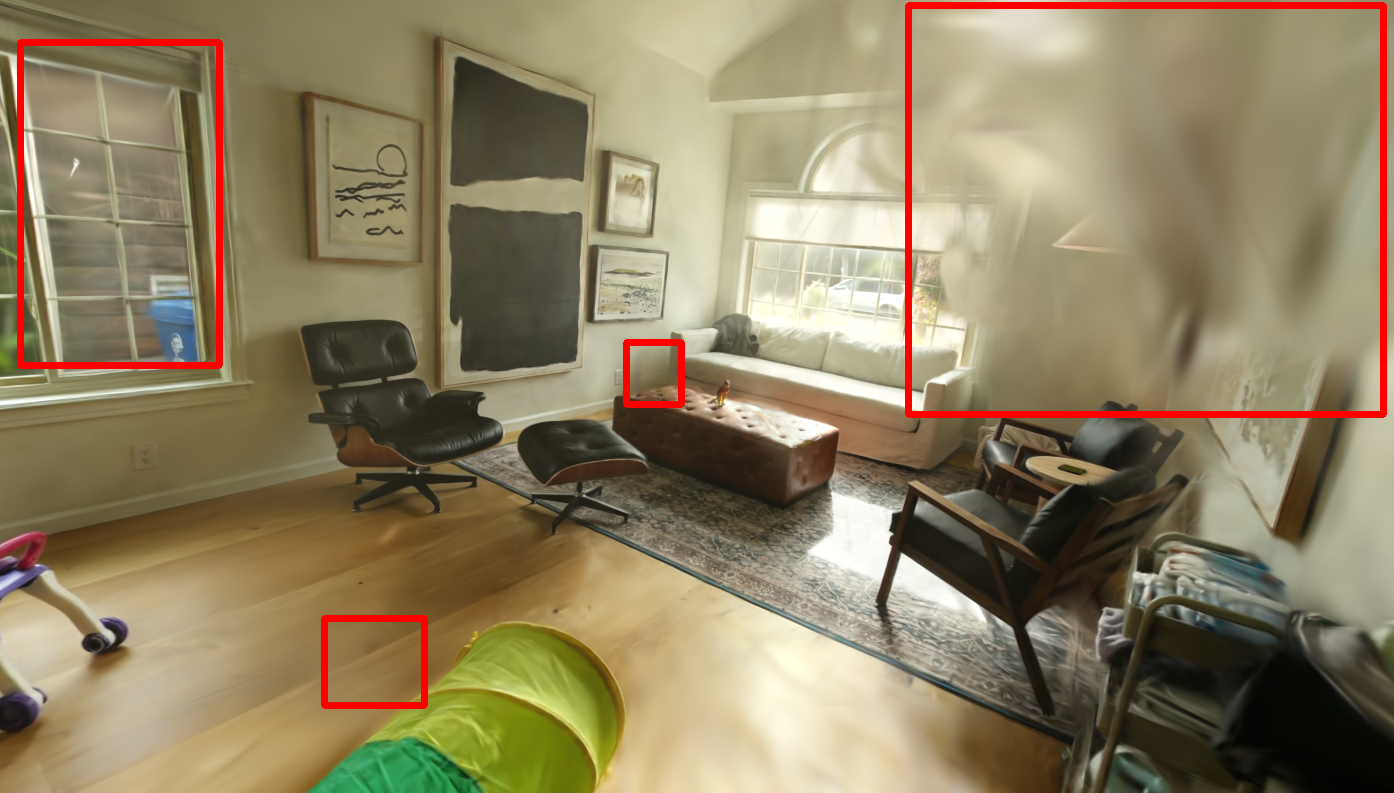}
\subcaption*{w/o NeRF Supervision}
\end{subfigure}
\subcaption{Qualitative Ablation Study of the NeRF-based Supervision.}\label{subfig:ablation2}
\vspace{.2cm}
\end{subfigure}
\begin{subfigure}{1.0\linewidth}
    \centering
    \resizebox{1.\textwidth}{!}{
   {\footnotesize
    \begin{tabular}{l|ccccc}
    \multicolumn{1}{c|}{} & SSIM$\uparrow$  & PSNR$\uparrow$ & LPIPS$\downarrow$ & \#G (M)$\downarrow$
    \\\hline
    \textbf{Ours} & 0.839 & 26.17 & 0.364 & 2.022 \\
     w/o NeRF Inititialization & 0.830 & 25.71 & 0.382 & 1.583 \\
     w/o NeRF Supervision & 0.835 & 25.79 & 0.372 & 1.849 \\
     w/o Pruning & 0.839 & 26.14 & 0.364 & 3.049 \\
    \hline
    \end{tabular}
     }
     }
     \subcaption{Quantitative Ablation Study on the \zipnerf{} Dataset.}
     \label{subfig:ablation3}
\end{subfigure}
\vspace{-.5cm}
\caption{
\textbf{Ablation Study.}
Without (w/o) the NeRF initialization, geometric and texture details might get lost (\ref{subfig:ablation1}). Without the NeRF supervision, floating artifacts appear if the views exhibit lighting or exposure changes (\ref{subfig:ablation2}). W/o pruning, the number of Gaussians is 1.5$\times$ larger without any quality improvements (\ref{subfig:ablation3}).
}
\label{fig:ablation}
\end{figure}

\boldparagraph{NeRF-based Initialization} The NeRF-based initialization leads to better quantitative and qualitative results (see~\figref{fig:ablation}).
In particular, smaller geometric and texture details might get lost, such as the back of the chair, the books behind the monitor, or the sticky notes on the wall in~\figref{subfig:ablation1}.

\boldparagraph{NeRF-based Supervision} The NeRF-based supervision leads to improved results compared to optimizing the scene representation on the input views directly. In particular, for scenes where the input views exhibit exposure or lighting variations, floating artifacts are introduced to model these effects as shown in~\figref{subfig:ablation2}.
In contrast, our strategy to optimize wrt.\ the NeRF-based supervision is more stable and leads to better reconstructions for real-world captures. 

\begin{figure}
\centering
\includegraphics[width=1.\linewidth,trim={0 1.4cm 0 0},clip]{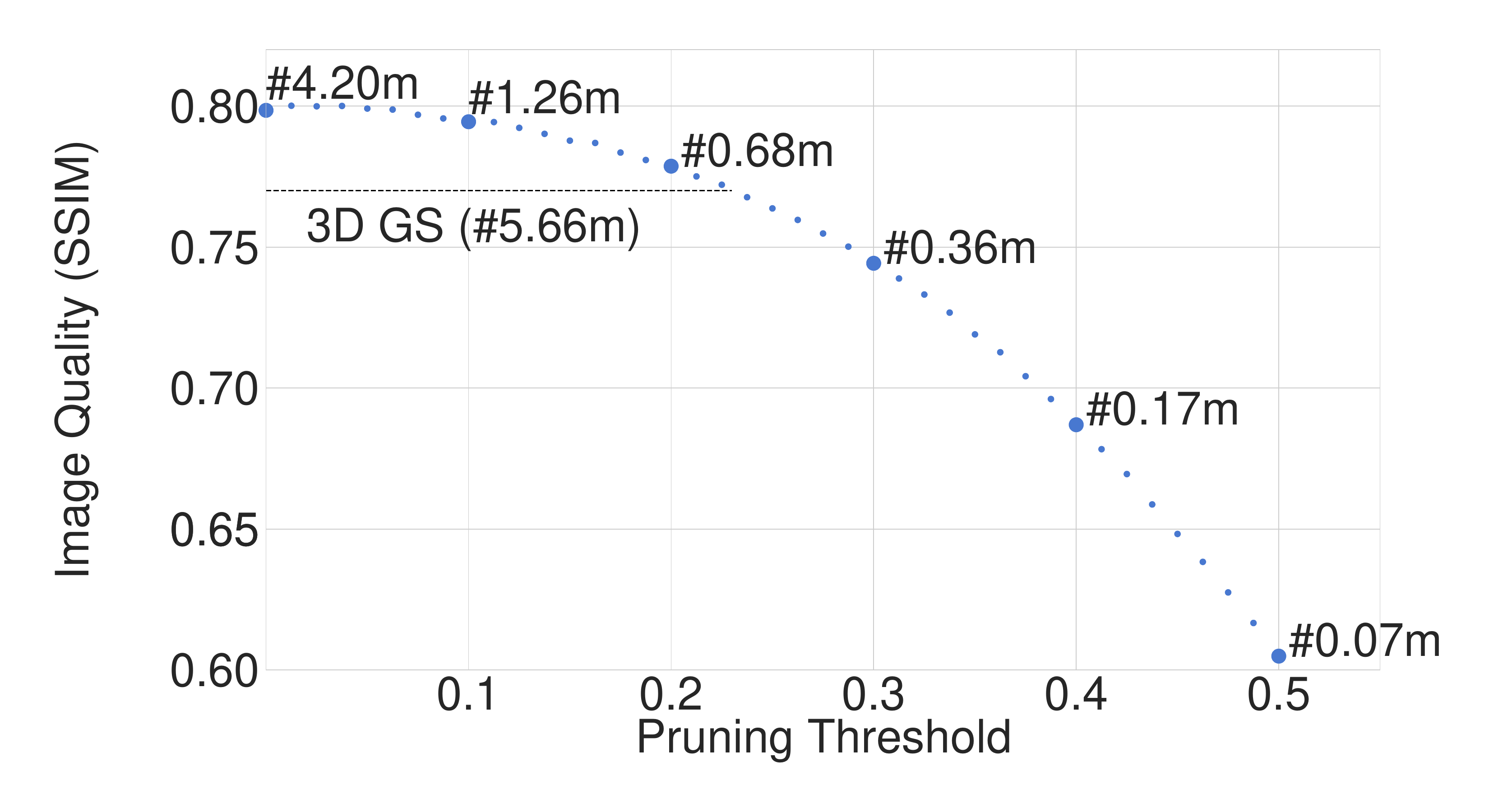}
\vspace{-.5cm}
\caption{
\textbf{Pruning.}
Pruning thresholds below $0.1$ maintain quality while reducing the point count by 4$\times$ (shown here on the mip-NeRF 360 Bicycle scene.). We find we can match the 3DGS quality (0.77) with a 10$\times$ reduction in Gaussians (5.66m vs.\ 0.59m).
}\label{fig:pruning}
\vspace{-.2cm}
\end{figure}

\begin{figure}[ht]
\centering
\includegraphics[width=1.\linewidth]{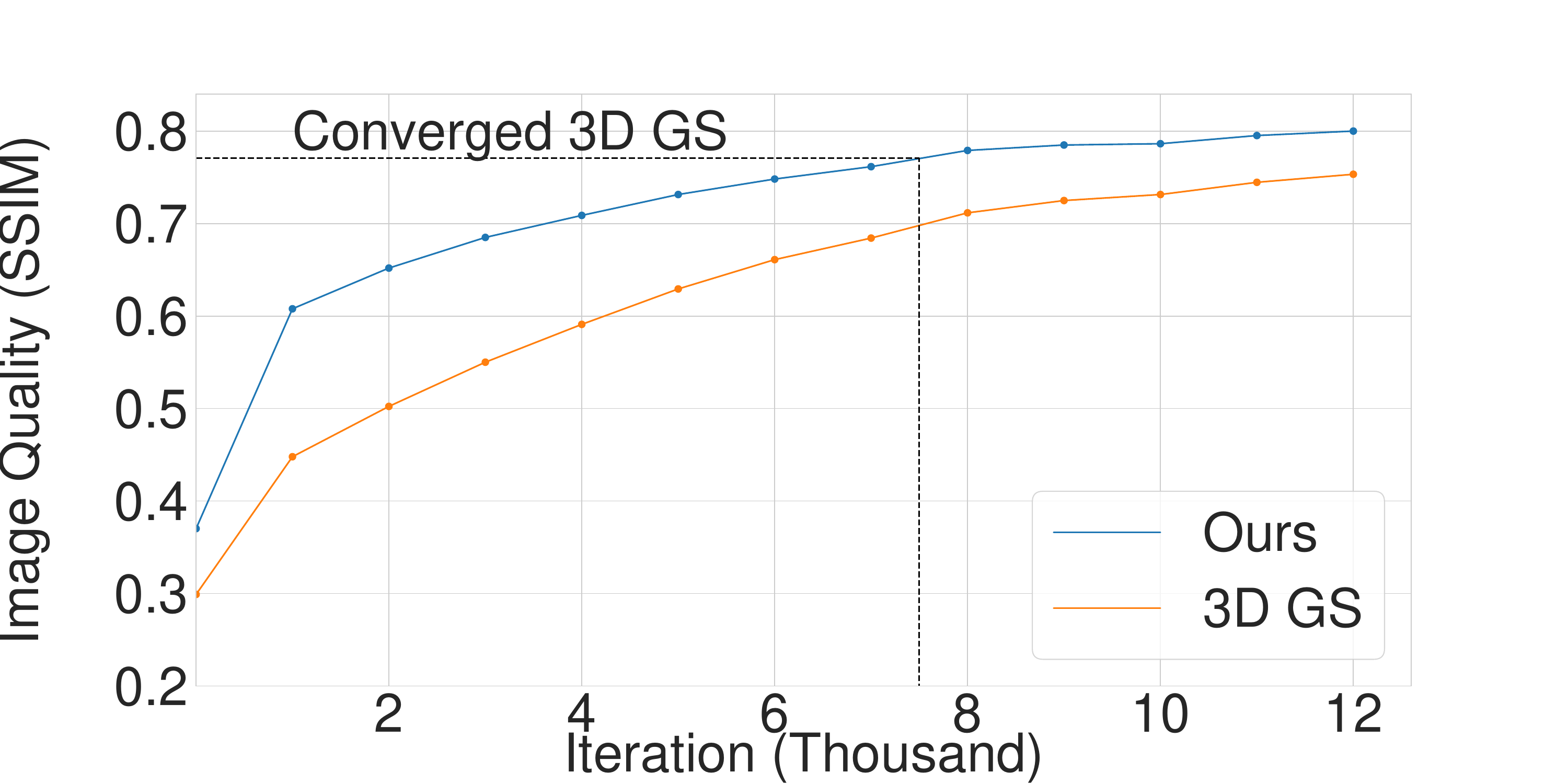}
\vspace{-.5cm}
\caption{
\textbf{Quality Progression.}
We compare the optimization progression of 3DGS and our default model on the mip-NeRF 360 Bicycle scene. We observe a steeper incline in SSIM and can match the final quality of 3DGS after less than 8k steps.
}\label{fig:optim}
\vspace{-.1cm}
\end{figure}

\boldparagraph{Pruning} In~\tabref{subfig:ablation3}, we find that not pruning leads to a small performance drop while exhibiting a significantly larger point count.
We hypothesize that a small pruning threshold removes redundancy leading to better generalization.
In summary, our pruning technique enables more compact scene representations  while maintaining high quality.
In~\figref{fig:pruning} we show that we can match the quality of 3DGS, despite having roughly 10$\times$  less Gaussians in the scene.
In~\figref{fig:optim}, we observe a faster increase in SSIM over the first iterations such that we can match the final 3DGS quality after only 8k iterations. 

\begin{table}[ht]
   \footnotesize
   \resizebox{1.0\textwidth}{!}{
    \begin{tabular}{l|cccc|ccccc}
        & \multicolumn{4}{c|}{mip-NeRF 360 dataset} & \multicolumn{4}{c}{ZipNeRF dataset} \\
        \multicolumn{1}{c|}{} & SSIM$\uparrow$  & PSNR$\uparrow$ & LPIPS$\downarrow$ & FPS$\uparrow$ & SSIM$\uparrow$  & PSNR$\uparrow$ & LPIPS$\downarrow$ & FPS$\uparrow$
        \\\hline
         \textbf{Ours} &  0.843 & 28.14 &  0.171 & 410 & 0.839 & 26.17 & 0.364 & 630 \\
        w/o Vis.\ Fil.\ &  0.843 & 28.14 &  0.171 & 373  & 0.839 & 26.17 & 0.364 & 435 \\
        \hline
         \textbf{Ours Light} & 0.826 & 27.56 & 0.213 & 907 & 0.838 &  26.11 &  0.368 &  748 \\
        w/o Vis.\ Fil.\ & 0.826 & 27.56 & 0.213 & 887 & 0.838 &  26.11 &  0.368 &  607 \\
        \hline
        \end{tabular}
        }
    \caption{
    \textbf{Visibility Filtering.}
    With this postprocessing, we achieve up to 10\% FPS increase on the mip-NeRF 360 scenes and up to 45\% improvement in rendering speed when scaling to the larger-scale ZipNeRF scenes while keeping the quality fixed.
    }\label{tab:visibility}
\end{table}

\boldparagraph{Visibility List-Based Rendering} Our visibility list-based rendering enables up to 10\% mean FPS speed up on the central object-focused MipNeRF360 scenes and a up to 45\% FPS increase on the larger house and apartment-level ZipNeRF scenes (see~\tabref{tab:visibility}). We conclude that this post-processing step is in particular important when scaling to more complex larger-scale scenes.

\boldparagraph{Limitations} 
Despite outperforming prior real-time methods, we observe a small gap to ZipNeRF on large-scale scenes which we aim to investigate in the future.

\section{Conclusion}

We presented RadSplat, a method combining the strengths of neural fields and point-based representations for robust real-time rendering of complex scenes. %
Using radiance fields as a prior and supervision signal leads to improved results and more stable optimization of point-based 3DGS representations.
Our novel pruning leads to more compact scenes with a significantly smaller scene size, whilst improving quality.
Finally, our novel test-time filtering further improves rendering speed without a quality drop.
Our method achieves state-of-the-art on common benchmarks while rendering up to 3,000$\times$ faster than prior works.

{
    \small
    \bibliographystyle{ieeenat_fullname}
    \bibliography{main}
}

\end{document}